\documentclass[lettersize,journal]{IEEEtran}
\usepackage{amsmath,amsfonts}
\usepackage{algorithmic}
\usepackage{algorithm}
\usepackage{array}
\usepackage[table]{xcolor}
\usepackage[caption=false,font=normalsize,labelfont=sf,textfont=sf]{subfig}
\usepackage{textcomp}
\usepackage{stfloats}
\usepackage{url}
\usepackage{verbatim}
\usepackage{graphicx}
\usepackage{cite}
\usepackage{textcomp}
\usepackage{stfloats}
\usepackage{url}
\usepackage{verbatim}
\usepackage{graphicx}
\usepackage{cite}
\usepackage{multicol}
\usepackage{multirow}
\usepackage{xcolor}

\usepackage{booktabs}

\usepackage{hyperref}
\hypersetup{
    colorlinks=true,
    citecolor=green, 
}

\begin{document}

\title{End-to-End Unmixing with Material Prompts for Hyperspectral Object Tracking}



\author{Xu Han, Mohammad Aminul Islam, Lei Wang, Zekun Long,  Guanmanyi Fu, \\Wangshu Cai, Kuldip K. Paliwal, Jun Zhou
\thanks{This work was supported by the Australian Research Council (ARC) Discovery Project DP250102625, ``Spectral-spatial-temporal object tracking in hyperspectral videos''. (Corresponding authors: Jun Zhou and Mohammad Aminul Islam)}

\thanks{ Xu Han, Zekun Long, Wangshu Cai,  and Jun Zhou are with the School of Information and Communication Technology, Griffith University, 
Australia (e-mail: xu.han2@griffithuni.edu.au, zekun.long@griffithuni.edu.au, wangshu.cai@griffithuni.edu.au, jun.zhou@griffith.edu.au).}
\thanks{Mohammad Aminul Islam is with the School of Information and Communication Technology, Griffith University, Australia and the Department of Computer Science and Mathematics, Bangladesh Agricultural University, Mymensingh 2202, Bangladesh  (e-mail: mohammadaminul.islam@griffith.edu.au).}
\thanks{Lei Wang and Kuldip K. Paliwal are with the School of Engineering and Built Environment, Griffith University, 
Australia (e-mail: l.wang4@griffith.edu.au, k.paliwal@griffith.edu.au).}
\thanks{Guanmanyi Fu is with the School of Environment and Science, Griffith University, 
Australia (e-mail: guanyiman.fu@griffith.edu.au).}

}

\markboth{Research report}
{Shell \MakeLowercase{\textit{et al.}}:Research report}


\maketitle

\begin{abstract}
Hyperspectral imagery encodes rich material properties that can improve tracking robustness under appearance ambiguity, illumination change, and background clutter. However, due to the limited availability of hyperspectral video data, many existing methods adapt pretrained RGB trackers via spatial or channel fusion strategies, largely neglecting the intrinsic material information in hyperspectral imagery. Moreover, the few material-aware approaches typically rely on external spectral unmixing pipelines that are decoupled from the tracking objective, limiting effective optimization of material representations for target localization. To address these limitations, we formulate hyperspectral object tracking as a joint optimization problem of material decomposition and target localization, coupling the two tasks via a weighted target-oriented unmixing loss that explicitly aligns material representations with localization accuracy. Specifically, we propose a material representation decomposition module for deep learning-based spectral unmixing with adaptive frequency decomposition. Building on the decomposed material representations, we further introduce a dual-branch wavelet-enhanced material prompt module that learns low- and high-frequency material prompts through efficient spatial-material interactions in the frequency domain. The framework is model-agnostic and can be seamlessly generalized to different unmixing backbones. Extensive experiments on standard hyperspectral tracking benchmarks demonstrate state-of-the-art performance and validate the effectiveness of the proposed end-to-end material-aware tracking framework. \href{https://github.com/han030927/E2EMPT}{[\textbf{Code}]}

\end{abstract}


\section{Introduction}
\begin{figure}[t]
    \centering
    \includegraphics[width=\linewidth]{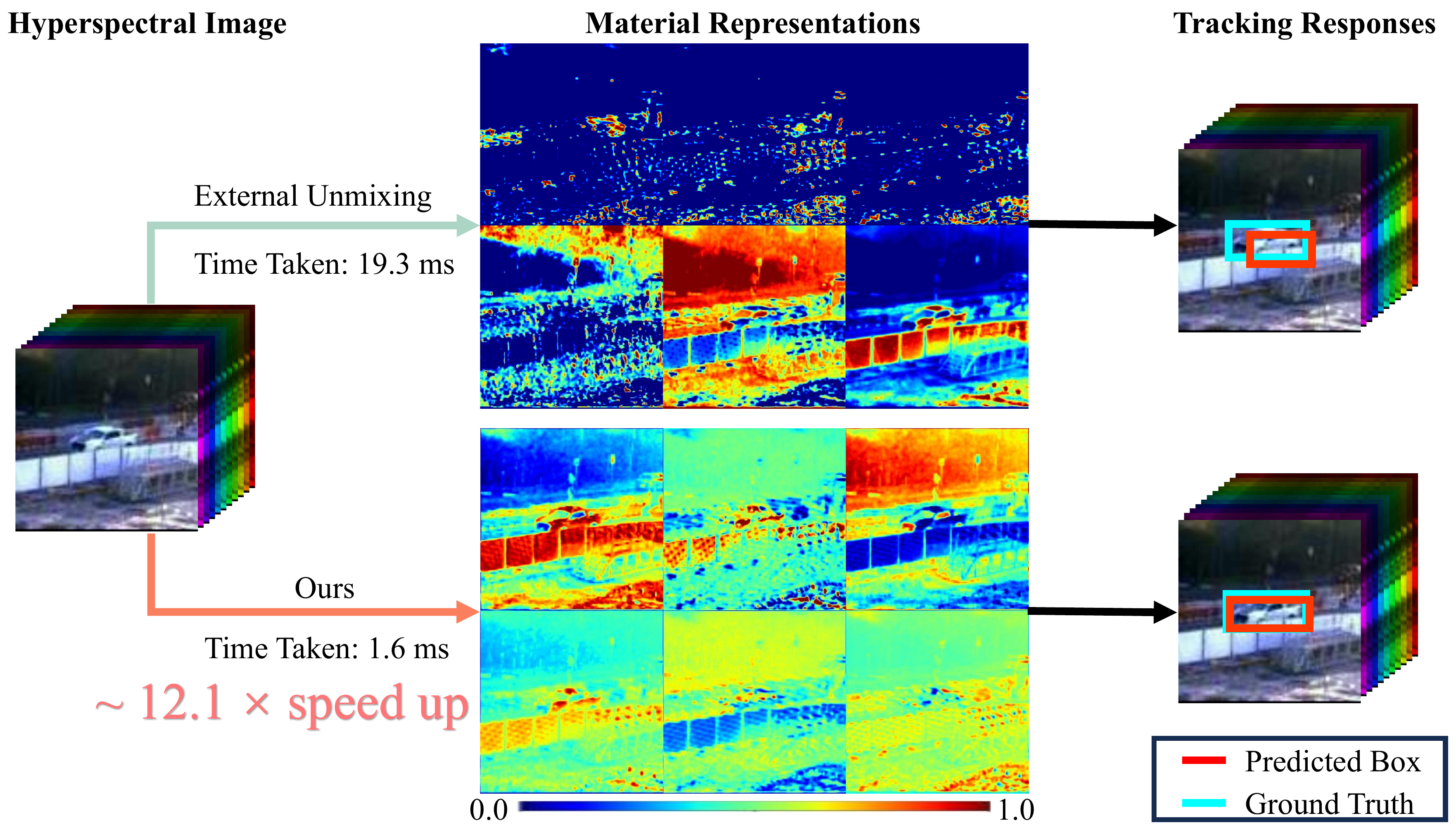}
    \caption{
Comparison between external unmixing and our Material Representation Decomposition Module (MRDM) for 
hyperspectral object tracking.
External unmixing introduces substantially higher processing latency in the current pipeline and produces material representations with more noise and redundancy, resulting in less concentrated tracking responses.
In contrast, our MRDM generates cleaner and more target-relevant material representations with much lower runtime, leading to more accurate target localization.
}
    \label{fig:abundance}
    \vspace{-12pt}
\end{figure}

\IEEEPARstart{O}{bject} tracking remains one of the most fundamental problems in computer vision, yet it remains a difficult task to achieve robust object localization under challenging conditions such as background clutter, illumination variation, and low spatial resolution~\cite{bai2024artrackv2,cao2023towards, chen2023seqtrack, islam2024hy,raj2025tracknetv4}. Over the past decade, tracking frameworks have evolved substantially, progressing from correlation filter methods~\cite{henriques2012exploiting,valmadre2017end} to Siamese networks~\cite{cen1018fully,li2019siamrpn++}, and more recently, transformer-based models~\cite{ye2022joint,bai2024artrackv2,
xie2024autoregressive}. Despite these advances, RGB-based trackers are fundamentally constrained by the limited information encoded in only three spectral bands. This has motivated multimodal tracking frameworks such as RGB-T~\cite{zhang2019multi}, RGB-E~\cite{NEURIPS2024_a29f0fc2}, and RGB-D~\cite{zhu2023rgbd1k}, which introduce complementary physical cues beyond conventional appearance information. Although these modalities provide additional thermal or depth information, they still lack the dense spectral measurements required to characterize intrinsic material properties. Consequently, their ability to exploit material-level discrimination for robust target representation and localization remains limited.

Hyperspectral imaging offers a promising alternative by offering not only spatial and temporal information but also dense spectral measurements that encode intrinsic material properties~\cite{islam2025ubstrack}. These material cues offer a distinctive advantage over RGB and conventional RGB+X modalities, especially in challenging scenarios where appearance information alone is insufficient for reliable target discrimination. Motivated by this capability, hyperspectral object tracking has attracted increasing attention, evolving from handcrafted 
approaches~\cite{xiong2025spatial,hou2022spatial} to deep learning frameworks~\cite{li2023siambag,chen2023spirit,li2026CCStrack}. However, annotating hyperspectral videos is labor-intensive and costly, resulting in limited training data. To alleviate this issue, existing methods have primarily followed two research directions. The first direction focuses on band selection strategies~\cite{li2020bae,islam2023background,li2023learning}, which select informative spectral subsets to facilitate knowledge transfer from pretrained RGB tracking models. The second direction explores visual-prompt-based methods~\cite{he2025hyperspectral, li2025multi,chen2025ssttrack}, which adapt frozen foundation models via lightweight trainable parameters. These approaches typically use false-color images as the primary backbone input while treating raw hyperspectral data as an auxiliary source of spatial, temporal, or spectral prompts. Moreover, multimodal fusion approaches~\cite{li2024material,zhao2022tftn,wang2025hyperspectral} have also been investigated by combining hyperspectral data with complementary modalities such as RGB or false-color imagery. Among them, MMF-Net~\cite{li2024material} further incorporates material representations through abundance fusion.

Despite recent progress, existing hyperspectral object tracking methods still suffer from two fundamental limitations. First, visual-prompt-based methods~\cite{he2025hyperspectral,li2025multi,xie2023vp} primarily treat hyperspectral spectra as an auxiliary cue while largely neglecting the intrinsic material properties encoded within the spectral signatures. As a result, the generated prompts are often dominated by background-related or spectrally redundant responses, particularly under mixed pixels and low spatial resolution, leading to weak target discriminability and unstable localization. Second, while material-aware methods~\cite{li2022material,li2024material} use hyperspectral unmixing to estimate abundance representations, the unmixing process is typically performed through external and decoupled unmixing pipelines optimized for spectral reconstruction rather than target 
localization. Under mixed pixels and low spatial resolution, such externally produced abundance maps are frequently dominated by background 
materials or contaminated by noisy responses, making them insufficiently discriminative for object tracking. Furthermore, these pipelines introduce extra computational overhead and processing latency in practical deployment. As shown in Fig.~\ref{fig:abundance}, the tracking responses produced by such decoupled pipelines are poorly concentrated in the target region, directly limiting the localization accuracy and robustness.

To overcome these limitations, we propose E2E-MPT, the first end-to-end framework to tightly integrate hyperspectral unmixing with object tracking under a joint optimization objective. Unlike existing approaches that treat unmixing as an independent preprocessing step, E2E-MPT unifies material decomposition and target localization, enabling mutual reinforcement. The proposed framework comprises three key components: a Material Representation Decomposition Module (MRDM), a Dual-Branch Wavelet-enhanced Material Prompt Module (DWMPM), and a Frequency Prompt Fusion Module (FPFM). Specifically, the MRDM introduces a plug-and-play integration mechanism that enables a deep learning-based unmixing backbone to be jointly optimized within the tracking framework. To enhance material discriminability, the MRDM further performs wavelet decomposition on abundance representations, disentangling target-relevant material cues from background interference into complementary low- and high-frequency branches. 

Building upon the decomposed material representations, the DWMPM jointly models hyperspectral patch tokens and material representations within each frequency branch. It combines cross-attention for global spectral-spatial interaction and convolution for local detail preservation, thereby generating discriminative material-aware prompts for the tracking backbone. Subsequently, the FPFM fuses the branch-wise prompt representations across multiple spatial levels to enhance feature integration and robustness. Furthermore, we introduce a weighted target-oriented unmixing loss to align the decomposition objective with localization accuracy. By assigning higher importance to target-associated spectral components during training, the proposed loss encourages target-discriminative material representations while suppressing background-dominated spectral responses.
The main \textbf{contributions} of this work are summarized as follows:
\begin{enumerate}
    \item We propose E2E-MPT, the first end-to-end framework that jointly optimizes hyperspectral unmixing and object tracking. It integrates decomposition and localization under a unified training objective, enabling material representations to directly facilitate target discrimination and localization.
    
    \item We design the MRDM and FPFM to decompose abundance representations into complementary frequency components 
    and fuse the resulting branch-wise material prompts, enabling the framework to exploit target-relevant material cues across multiple spectral scales while suppressing redundancy and background responses.
    
    \item We develop the DWMPM to jointly model hyperspectral tokens and material representations within each frequency branch through a combination of cross-attention and convolutional operations, achieving effective global spectral-spatial interaction while preserving local structural details.
    
    \item We introduce a weighted target-oriented unmixing loss that explicitly aligns hyperspectral unmixing with the tracking objective, 
    encouraging target-discriminative material decomposition and suppressing background-dominated spectral responses.

    \item Extensive experiments on HOTC2020, HOTC2023, and HOTC2024 demonstrate that the proposed E2E-MPT framework generalizes across multiple deep learning-based unmixing backbones and achieves state-of-the-art tracking performance.
\end{enumerate}

The remainder of this paper is organized as follows. Section~\ref{sec:related} reviews related work on hyperspectral tracking, vision prompt learning, and hyperspectral unmixing. Section~\ref{sec:method} introduces the proposed E2E-MPT framework in detail. Section~\ref{sec:experiments} presents the experimental results and comprehensive analysis on benchmark datasets. Finally, Section~\ref{sec:conclusion} concludes the paper.

\begin{figure*}
\vspace{-0.6cm}
    \centering
    \includegraphics[width=0.95\textwidth]{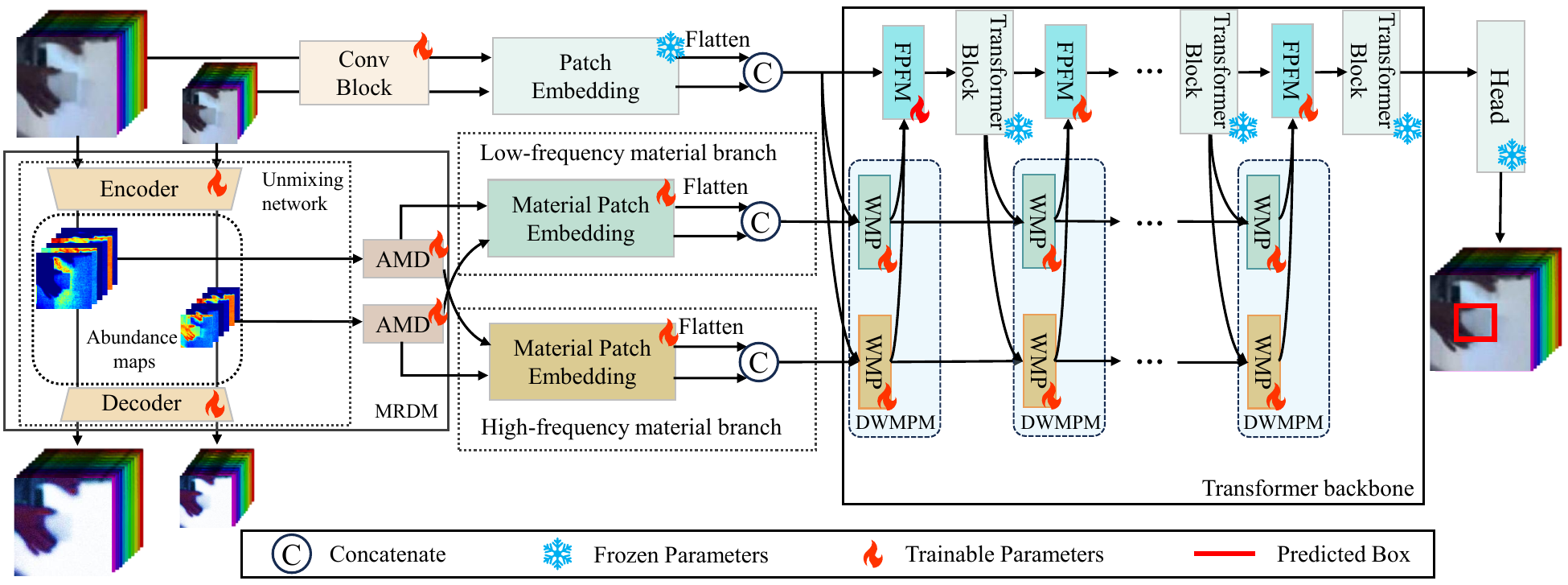}
    \vspace{-0.2cm}
    \caption{
Overview of the proposed E2E-MPT framework. The backbone branch processes hyperspectral template and search images using a frozen tracking backbone. In parallel, the material branch uses the Material Representation Decomposition Module (MRDM), consisting of an autoencoder-based unmixing network and an Abundance Map Decomposition (AMD) module, to extract and decompose abundance maps into low- and high-frequency material representations. Subsequently, these representations are transformed into material prompts by the Dual-Branch Wavelet-enhanced Material Prompt Module (DWMPM), where each DWMPM contains two Wavelet-enhanced Material Prompt (WMP) blocks corresponding to the two frequency branches. Finally, the generated prompts are fused via the Frequency Prompt Fusion Module (FPFM) and injected into multiple transformer blocks for material-aware target localization.
}
    \label{fig:framework}
    \vspace{-0.6cm}
\end{figure*}

\section{Related Work}
\label{sec:related}

\noindent\textbf{Hyperspectral tracking.}
Existing hyperspectral tracking methods fall into two broad categories: hand-crafted feature-based trackers~\cite{qian2018object,xiong2020material,xiong2025spatial} and deep learning-based trackers~\cite{HDSP,li2020bae,islam2023multi}. Hand-crafted approaches combine correlation filters with spectral-spatial descriptors, with some incorporating unmixing cues for target representation~\cite{xiong2020material}. While effective in constrained settings, their limited representational capacity degrades tracking performance under complex scenarios. Deep learning-based methods address this issue by transferring RGB-pretrained backbones to the hyperspectral domain~\cite{HDSP,wu2024domain,chen2024sense}. More recently, prompt-based strategies have been introduced to adapt foundation trackers to hyperspectral inputs~\cite{HDSP,li2025multi,chen2024phtrack}. However, existing prompt designs operate at the feature level and are not grounded in abundance-level material composition. When unmixing is used, it remains decoupled from the tracking objective, producing representations optimized for spectral reconstruction rather than target discrimination and localization. As a consequence, the generated prompts are dominated by background responses under mixed pixels and low spatial resolution. In contrast, our method produces prompts from abundance representations that are jointly optimized with the tracking objective, directly aligning material decomposition with target localization.

\noindent\textbf{Vision prompt learning.}
Prompt learning was first introduced in NLP and later brought to vision tasks~\cite{jia2022visual}, where a small number of trainable parameters are introduced to adapt large pretrained models to downstream tasks without full fine-tuning. In object tracking, ViPT~\cite{zhu2023visual} learns modality-aware prompts to adapt a foundation tracker across multimodal inputs, while PiVOT~\cite{chen2025improving} generates instance-aware prompts via CLIP~\cite{radford2021learning} for distractor suppression. In hyperspectral tracking, 
DaSSP-Net\cite{li2025multi} uses a domain adaptor with spatial-spectral prompt learning to unify tracking across hyperspectral modalities. Unlike these methods that derive prompts from spatial or spectral features, our approach generates material-aware prompts from abundance representations that are jointly optimized with the tracking objective. This enables prompt learning to be grounded in intrinsic material composition rather than appearance-level representations, thereby improving target discriminability and localization robustness under complex tracking scenarios.

\noindent\textbf{Hyperspectral unmixing.}
Hyperspectral unmixing aims to decompose mixed pixels into constituent endmember spectra and their corresponding abundance maps. The derived abundance maps encode the spatial distribution and mixture proportions of materials in the scene. Deep learning has significantly advanced this field through autoencoder-based frameworks~\cite{palsson2018hyperspectral,palsson2020convolutional}, where the encoder maps spectral inputs to abundance representations and the decoder reconstructs the hyperspectral signal while learning endmember features. Extensions incorporating multi-scale spatial modeling~\cite{yu2022multi} have further improved performance in complex 
scenes. Beyond reconstruction, recent works have demonstrated that abundance representations serve as physically meaningful intermediate features for downstream vision tasks such as boundary detection~\cite{Al2021boundary}, image synthesis~\cite{yu2024unmixdiff} and salient object detection~\cite{liang2018material} suggesting their broader utility as task-agnostic material descriptors. However, no existing method has integrated unmixing into object tracking through joint end-to-end optimization. In previous works, abundance representations are always produced externally and optimized for spectral reconstruction rather than target discrimination. E2E-MPT addresses this gap by jointly optimizing material decomposition and target localization, enabling abundance representations to directly serve object tracking rather than a decoupled preprocessing step.

\section{End-to-End Material Prompt Tracking}
\label{sec:method}

\noindent\textbf{Overview.} Fig.~\ref{fig:framework} illustrates the proposed End-to-End Material Prompt Tracking (E2E-MPT) framework. Given a hyperspectral template-search pair as input, E2E-MPT jointly performs material decomposition and target localization through three key components: (i) the Material Representation Decomposition Module (MRDM), which generates and decomposes abundance representations into complementary low- and high-frequency branches; (ii) the Dual-Branch Wavelet-enhanced Material Prompt Module (DWMPM), which produces material-aware prompts within each frequency branch; and (iii) the Frequency Prompt Fusion Module (FPFM), which adaptively fuses branch-wise prompts and injects them into the tracking backbone for target localization. We next introduce the problem formulation.

\noindent\textbf{Problem setup.}
Given a hyperspectral video with an initial target bounding box $\mathbf{b}_0$, the goal is to learn a tracker $\mathrm{\mathbf{T}}^{\mathrm{HS}}: \{ \mathbf{X}^{\mathrm{HS}}, \mathbf{b}_0 \} \!\rightarrow\! \mathbf{b}$ that predicts the target bounding box $\mathbf{b}$ in subsequent frames, where $\mathbf{X}^{\mathrm{HS}} \!\in\! \mathbb{R}^{n \times h \times w}$ denotes a hyperspectral frame with $n$ spectral bands and spatial size $h \!\times\! w$.
%
We adopt the linear mixture model for hyperspectral unmixing, which decomposes a hyperspectral image $\mathbf{X}^{\mathrm{HS}} \!\in\! \mathbb{R}^{n \times p}$ ($p \!=\! h \!\times\! w$) as
\begin{equation}
\label{eq:unmixing_definition}
\mathbf{X}^{\mathrm{HS}} = \mathbf{M}\mathbf{A} + \mathbf{E},
\end{equation}
where $\mathbf{M} \!\in\! \mathbb{R}^{n \times r}$ denotes the endmember matrix, $r$ denotes the number of endmembers,  $\mathbf{A} \!\in\! \mathbb{R}^{r \times p}$ denotes the abundance maps satisfying non-negativity and sum-to-one constraints~\cite{su2017nonnegative}, and $\mathbf{E} \!\in\! \mathbb{R}^{n 
\times p}$ is the residual noise. E2E-MPT jointly optimizes both problems under a unified objective, enabling abundance maps $\mathbf{A}$ to serve as target-discriminative material representations rather than reconstruction-oriented spectral decompositions.
We next describe the key modules.

\subsection{Material Representation Decomposition Module}
\label{subsec:mrdm}

\noindent\textbf{Unmixing network.}
To explicitly characterize pixel-wise material composition, E2E-MPT incorporates a deep autoencoder-based unmixing network. 
The unmixing process is parameterized as
\begin{equation}
\label{eq:unmixing_network}
\mathbf{A} = \mathrm{Encoder}(\mathbf{X}), \qquad
\hat{\mathbf{X}} = \mathrm{Decoder}(\mathbf{A}),
\end{equation}
where the encoder maps the hyperspectral input $\mathbf{X}\in\mathbb{R}^{n\times h\times w}$ to the abundance representation $\mathbf{A}\in\mathbb{R}^{r\times h\times w}$, and the decoder reconstructs the spectral response $\hat{\mathbf{X}}\in\mathbb{R}^{n\times h\times w}$. The framework interfaces with the unmixing network solely through $\mathbf{A}$, making it compatible with existing 
autoencoder-based unmixing architecture, including DAEU~\cite{palsson2018hyperspectral} and CNNAEU~\cite{palsson2020convolutional}, without modification to the tracking pipeline.

\noindent\textbf{Abundance map decomposition.}
\begin{figure}[tbp]
    \centering
    \includegraphics[height=0.9\linewidth]{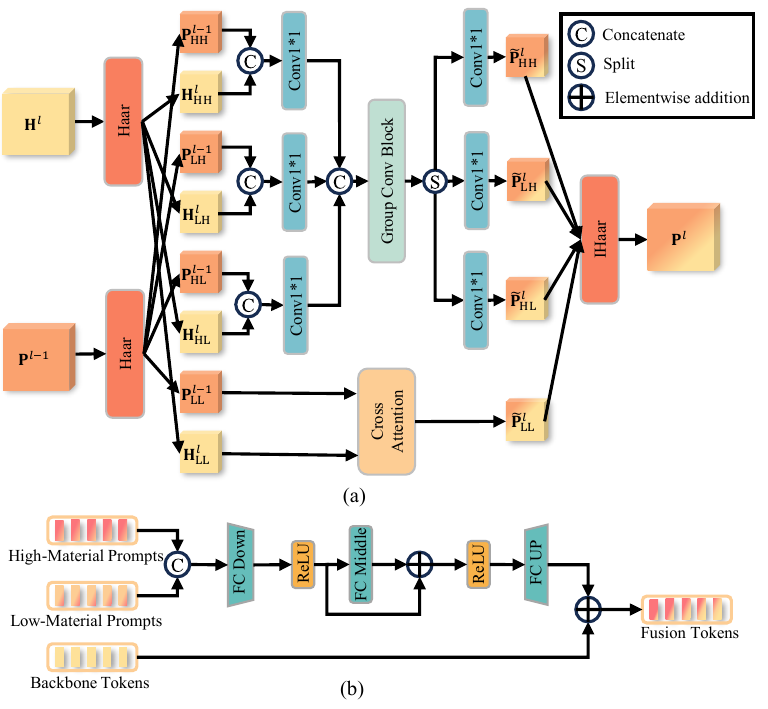}
    \caption{
   Architecture of the proposed (a) Wavelet-enhanced Material Prompt (WMP) block and (b) Frequency Prompt Fusion Module (FPFM).
    }
    \label{fig:prompt_block}
    \vspace{-0.3cm}
\end{figure}
Raw abundance representations present two challenges for direct use as tracking prompts. First, their channel semantics are defined in the unmixing space and are misaligned with the feature space of the tracking backbone. Second, each abundance map entangles complementary material cues-dominant shared responses that provide stable target support, and subtle inter-channel contrastive variations that distinguish the target from confusing background regions-whose direct injection into the tracker introduces redundancy and unstable interference.

To address both issues, we design an Abundance Map Decomposition (AMD) module. AMD first applies a $1\times1$ channel adaptor to transform the original  endmember-wise coefficients into a tracking-conditioned material basis
\begin{equation}
\label{eq:amd_adaptor}
\mathbf{A}_{p}=\mathrm{Conv}_{1\times1}(\mathbf{A}),
\end{equation}
where $\mathbf{A}\!\in\!\mathbb{R}^{r\times h\times w}$ is the raw abundance map and $\mathbf{A}_{p}\!\in\!\mathbb{R}^{2c\times h\times w}$ is the adapted representation with $c$ channels per material branch. 
We set $c\!=\!3$ so that $\mathbf{A}_{p}\!\in\!\mathbb{R}^{6\times h\times w}$ and each branch has three channels before material patch embedding. This adaptor simultaneously aligns the material representation with the tracker feature space and learns a channel basis more suitable for subsequent frequency decomposition. 
A channel-wise 1D Haar wavelet transform is then applied to explicitly decouple the two complementary material cues:
\begin{equation}
\label{eq:1dhaar}
\begin{aligned}
\mathbf{M}_{L}[k,h,w] &=
\frac{1}{\sqrt{2}}\bigl(\mathbf{A}_{p}[2k-1,h,w]+
\mathbf{A}_{p}[2k,h,w]\bigr),\\
\mathbf{M}_{H}[k,h,w] &=
\frac{1}{\sqrt{2}}\bigl(\mathbf{A}_{p}[2k-1,h,w]-
\mathbf{A}_{p}[2k,h,w]\bigr),
\end{aligned}
\end{equation}
where $k\!\!=\!\!1,\dots,c$, and $\mathbf{M}_{L}, \mathbf{M}_{H}\!\!\in\!\! \mathbb{R}^{c\times h\times w}$ denote low- and high-frequency material branches, respectively. $\mathbf{M}_{L}$ aggregates dominant and smooth material responses to provide stable target support, while $\mathbf{M}_{H}$ isolates inter-channel contrastive variations to suppress background interference and enhance local discrimination. 
The two branches transform reconstruction-oriented abundance maps into complementary tracking-oriented material representations for prompt generation.
Below, we introduce the Dual-Branch Wavelet-enhanced Material Prompt Module (DWMPM).

\subsection{Dual-Branch Wavelet-enhanced Material Prompt Module}
\label{subsec:wmp}

DWMPM progressively refines material prompts via interactions with backbone features across transformer layers, operating on low- and high-frequency branches generated by MRDM through parallel Wavelet-enhanced Material Prompt (WMP) blocks. The asymmetric fusion design, using cross-attention for low-frequency components and lightweight convolution for high-frequency components, is motivated by their distinct structural properties: low-frequency components encode globally distributed material responses that benefit from long-range feature conditioning, whereas high-frequency components capture spatially local and fine-grained material variations that are more efficiently modeled by local convolutional operations.

The low- and high-frequency branches $\mathbf{M}_{L}$ and $\mathbf{M}_{H}$ are first projected into the backbone-compatible token space
\begin{equation}
\begin{aligned}
\mathbf{T}_{L}^{0}&=\mathrm{Flatten}(\mathrm{PatchEmbed}_{m}(\mathbf{M}_{L})), \\
\mathbf{T}_{H}^{0}&=\mathrm{Flatten}(\mathrm{PatchEmbed}_{m}(\mathbf{M}_{H})),
\end{aligned}
\end{equation}
where $\mathbf{T}_{L}^{0}$ and $\mathbf{T}_{H}^{0}$ denote the tokenized low- and high-frequency material prompts, reshaped into 2D feature maps $\mathbf{P}_{L}^{0}$ and $\mathbf{P}_{H}^{0}$ before entering each WMP block.

At layer $l$, two WMP blocks take the propagated branch-wise prompts $\mathbf{P}_{L}^{l-1}$ and $\mathbf{P}_{H}^{l-1}$ together with the backbone feature $\mathbf{H}^{l}$ as input, and output the updated prompts $\mathbf{P}_{L}^{l}$ and $\mathbf{P}_{H}^{l}$, respectively. The architecture is detailed in Fig.~\ref{fig:prompt_block}(a); the same formulation is applied independently to both branches.

For brevity, we describe a generic WMP block with input prompt $\mathbf{P}^{l-1}$ and backbone feature $\mathbf{H}^{l}$. We first apply a 2D Haar wavelet transform to both inputs:
\begin{align}
\left(
\mathbf{H}^{l}_{\mathrm{LL}},
\mathbf{H}^{l}_{\mathrm{HL}},
\mathbf{H}^{l}_{\mathrm{LH}},
\mathbf{H}^{l}_{\mathrm{HH}}
\right)
& =
\mathrm{Haar}(\mathbf{H}^{l}), \\
\left(
\mathbf{P}^{l-1}_{\mathrm{LL}},
\mathbf{P}^{l-1}_{\mathrm{HL}},
\mathbf{P}^{l-1}_{\mathrm{LH}},
\mathbf{P}^{l-1}_{\mathrm{HH}}
\right)
& =
\mathrm{Haar}(\mathbf{P}^{l-1}),
\end{align}
where $\mathrm{LL}$ denotes the low-frequency approximation component, while $\mathrm{HL}$, $\mathrm{LH}$, and $\mathrm{HH}$ correspond to the high-frequency detail components.
The low-frequency backbone and prompt components, $\mathbf{H}^{l}_{\mathrm{LL}}$ and $\mathbf{P}^{l-1}_{\mathrm{LL}}$, encode global structures and dominant material responses. These are fused via cross-attention to enable adaptive long-range conditioning of material representations on backbone features
\begin{equation}
\label{eq:LF_fusion}
\tilde{\mathbf{P}}^{l}_{\mathrm{LL}}
= \mathrm{CrossAttention}(\mathbf{P}^{l-1}_{\mathrm{LL}}, 
\mathbf{H}^{l}_{\mathrm{LL}}),
\end{equation}
where the cross-attention operator follows~\cite{li2025multi}.
The high-frequency components contain fine-grained material variations and spatially local discriminative cues. For each $i\in\{\mathrm{HL},\mathrm{LH},\mathrm{HH}\}$, we adopt a group convolutional fusion
\begin{equation}
\label{eq:HF_fusion}
\tilde{\mathbf{P}}^{l}_i
=
\mathrm{Conv}_{1\times1}^{(\mathrm{o})}
\!\Big(
\mathrm{GCB}
\!\big(
\mathrm{Conv}_{1\times1}^{(\mathrm{i})}
([\mathbf{H}^{l}_i \Vert \mathbf{P}^{l-1}_i])
\big)
\Big),
\end{equation}
where $\mathrm{GCB}$ denotes a group convolution block comprising a $3\times3$ group convolution, Batch Normalization, and LeakyReLU activation, and $[\cdot \Vert \cdot]$ denotes channel-wise concatenation. The input and output $1\times1$ convolutions serve as channel projections for computational efficiency. 

After fusing both low- and high-frequency components, an inverse Haar wavelet transform (IHaar) is applied to reconstruct the spatial structure,
\begin{equation}
\label{eq:IDWT}
\mathbf{P}^{l}
=
\mathrm{IHaar}
\big(
\tilde{\mathbf{P}}^{l}_{\mathrm{LL}},
\tilde{\mathbf{P}}^{l}_{\mathrm{HL}},
\tilde{\mathbf{P}}^{l}_{\mathrm{LH}},
\tilde{\mathbf{P}}^{l}_{\mathrm{HH}}
\big).
\end{equation}
Here, $\mathbf{P}^{l}$ denotes the branch-wise material prompt generated by one WMP block at layer $l$.
In our framework, the two WMP blocks inside each DWMPM respectively produce the low-frequency prompt $\mathbf{P}^{l}_{L}$ and the high-frequency prompt $\mathbf{P}^{l}_{H}$, which are further fused by the subsequent Frequency Prompt Fusion Module (FPFM) described below.

\subsection{Frequency Prompt Fusion Module}
FPFM adaptively fuses the low- and high-frequency branch prompts $\mathbf{P}^{l}_{L}$ and $\mathbf{P}^{l}_{H}$ in a compact latent space, as shown in Fig.~\ref{fig:prompt_block}(b). Instead of directly adding heterogeneous frequency components in the backbone feature space, FPFM projects them into a shared latent space where the model can learn an adaptive balance between dominant material support and contrastive material variations. The two branch prompts are first concatenated along the channel dimension
\begin{equation}
\label{eq:fpfm_concat}
\mathbf{P}^{l}_{c} = [\mathbf{P}^{l}_{L} \Vert \mathbf{P}^{l}_{H}],
\end{equation}
and projected to a reduced latent dimension via a down-projection layer
\begin{equation}
\label{eq:fpfm_down}
\mathbf{Z}^{l}_{0} = \phi\!\left(\mathrm{FC}_{\mathrm{down}}
(\mathbf{P}^{l}_{c})\right),
\end{equation}
where $\phi(\cdot)$ denotes LeakyReLU activation. A lightweight bottleneck with a residual connection then refines the fused representation while preserving individual branch information
\begin{equation}
\label{eq:fpfm_mid}
\mathbf{Z}^{l}_{1} = \mathrm{FC}_{\mathrm{middle}}(\mathbf{Z}^{l}_{0}) 
+ \mathbf{Z}^{l}_{0}.
\end{equation}

After that, an up-projection recovers the original token dimension to produce the fused material prompt $\mathbf{P}^{l}_{f}$
\begin{equation}
\label{eq:fpfm_up}
\mathbf{P}^{l}_{f} = \mathrm{FC}_{\mathrm{up}}\!\left(\phi
(\mathbf{Z}^{l}_{1})\right).
\end{equation}
Finally, the fused prompt is injected into the backbone via residual addition
\begin{equation}
\label{eq:fpfm_out}
\mathbf{H}^{l}_{p} = \mathbf{H}^{l} + \mathbf{P}^{l}_f,
\end{equation}
where $\mathbf{H}^{l}_{p}$ denotes the prompt-enhanced backbone representation at layer $l$. This design enables the tracking backbone 
to exploit complementary material cues from both frequency branches without disrupting the pretrained feature representations through 
direct feature overwriting.

\subsection{Prediction Module and Training Objective }
\label{subsec:Loss}
\noindent\textbf{Prediction head.}
Following OSTrack \cite{ye2022joint}, the prompt-enhanced search-region tokens are reshaped into a 2D spatial feature map and fed into a prediction head comprising three parallel branches: classification, offset regression, scale regression. Each branch consists of five stacked Conv-BN-ReLU layers with $3\times3$ kernels followed by a $1\times1$ output projection. The classification branch predicts the probability that each pixel belongs to the target foreground, the offset branch compensates for discretization errors introduced by the reduced spatial resolution, and the scale regression branch estimates target width and height.
%
The tracking objective
\begin{equation}
\label{eq:trackingloss}
\mathcal{L}_{\mathrm{tracking}}
=
\mathcal{L}_{\mathrm{cls}}
+
\lambda_{\mathrm{IoU}}\mathcal{L}_{\mathrm{IoU}}
+
\lambda_{\mathrm{L1}}\mathcal{L}_{\mathrm{L1}},
\end{equation}
where $\mathcal{L}_{\mathrm{cls}}$ denotes the weighted focal loss, and $\mathcal{L}_{\mathrm{L1}}$ and $\mathcal{L}_{\mathrm{IoU}}$ denote the $\ell_1$ and generalized IoU regression losses, with $\lambda_{\mathrm{IoU}}=2$ and $\lambda_{\mathrm{L1}}=5$ following \cite{ye2022joint}. 

\noindent\textbf{Target-oriented unmixing loss.}
To explicitly align hyperspectral unmixing with the tracking objective, we introduce a target-oriented unmixing loss that steers material decomposition toward target-relevant regions rather than optimizing uniformly over the entire scene.
In our implementation, we followed ~\cite{palsson2018hyperspectral} to adopt a simple reconstruction loss formed by spectral angle distance (SAD) and mean squared error (MSE),
\begin{equation}
\label{eq:reconstruction_loss}
\mathcal{L}_{\mathrm{rec}}
=
\mathcal{L}_{\mathrm{SAD}}(\hat{\mathbf{X}}, \mathbf{X})
+
\mathcal{L}_{\mathrm{MSE}}(\hat{\mathbf{X}}, \mathbf{X}),
\end{equation}
where $\mathcal{L}_{\mathrm{SAD}}$ encourages spectral consistency and $\mathcal{L}_{\mathrm{MSE}}$ constrains reconstruction fidelity.

To bias unmixing toward target-relevant material reconstruction, we construct a binary spatial mask $\mathbf{V}\in\{0,1\}^{h\times w}$ over the search region, derived from the candidate elimination scores of the tracking backbone, where $\mathbf{V}_{ij}=0$ indicates a  target-relevant pixel and $\mathbf{V}_{ij}=1$ indicates a background pixel. The target-oriented unmixing loss is then formulated as
\begin{equation}
\label{eq:target_unmixing_loss}
\begin{aligned}
&\mathcal{L}_{\mathrm{unmixing}}
\;=\;
\mathcal{L}_{\mathrm{rec}}
\bigl(\hat{\mathbf{X}}^{\mathrm{template}}, \mathbf{X}^{\mathrm{template}}\bigr)\\
&+ (1-\lambda_{\mathrm{ce}})
\mathcal{L}_{\mathrm{rec}}
\bigl((\mathbf{1}-\mathbf{V}) \odot \hat{\mathbf{X}}^{\mathrm{search}},
(\mathbf{1}-\mathbf{V}) \odot \mathbf{X}^{\mathrm{search}}\bigr),\\
&+ \lambda_{\mathrm{ce}}
\mathcal{L}_{\mathrm{rec}}
\bigl(\mathbf{V} \odot \hat{\mathbf{X}}^{\mathrm{search}},
\mathbf{V} \odot \mathbf{X}^{\mathrm{search}}\bigr) 
\end{aligned}
\end{equation}
where $\odot$ denotes element-wise multiplication and $\lambda_{\mathrm{ce}} \in(0,1)$ controls the relative weight between target-relevant and background regions.

\noindent\textbf{Overall objective.}
The framework is jointly optimized under a unified loss
\begin{equation}
\label{eq:total_loss}
\mathcal{L}_{\mathrm{total}}
=
(1-\lambda_u)\mathcal{L}_{\mathrm{tracking}}
+
\lambda_{u}\mathcal{L}_{\mathrm{unmixing}},
\end{equation}
where $\lambda_u$ balances the tracking and unmixing objectives during joint optimization. This formulation couples material decomposition with target localization during training: the tracking objective guides unmixing toward target-discriminative representations, while the unmixing objective provides structured material supervision that enriches the tracking backbone with physically grounded spectral cues.

\section{Experiments}
\label{sec:experiments}

\subsection{Experimental Setup}
\begin{table*}[tbp]
\centering
\small
\vspace{-0.3cm}
\renewcommand{\arraystretch}{1.15}
\setlength{\tabcolsep}{5.5pt}
\caption{Performance comparison on HOTC2020, HOTC2023 and HOTC2024. For RGB trackers, we use false-color images (FCI) or RGB images as input, while for hyperspectral trackers, we use hyperspectral images (HSI) as input.}
\vspace{-0.2cm}
\label{tab:hotc_results}
\resizebox{\linewidth}{!}{
\begin{tabular}{l c|cc | cc cc cc  | cc cc cc }
\hline
\multirow{3}{*}{Method} & \multirow{3}{*}{Venue}
& \multicolumn{2}{c|}{HOTC2020}
& \multicolumn{6}{c|}{HOTC2023}
& \multicolumn{6}{c}{HOTC2024} \\
&
& \multicolumn{2}{c|}{RGB/VIS}
& \multicolumn{2}{c}{FCI/NIR}
& \multicolumn{2}{c}{FCI/RedNIR}
& \multicolumn{2}{c|}{FCI/VIS}
& \multicolumn{2}{c}{FCI/NIR}
& \multicolumn{2}{c}{FCI/RedNIR}
& \multicolumn{2}{c}{FCI/VIS}

\\
& 
& DP & AUC & DP & AUC & DP & AUC 
& DP & AUC& DP & AUC & DP & AUC & DP & AUC \\
\hline
SeqTrack ~\cite{chen2023seqtrack}
& CVPR'23
&0.946&0.689 & 0.834 & 0.643 & 0.631 & 0.470 & 0.831 & 0.603
& 0.818 & 0.632 & 0.483 & 0.370 & 0.691 & 0.525 \\

TCTrack++ ~\cite{cao2023towards}
& TPAMI'23
&0.872&0.623 & 0.822 & 0.583 & 0.534 & 0.376 & 0.690 & 0.479
& 0.818 & 0.590 & 0.432 & 0.264 & 0.591 & 0.412\\

SMAT ~\cite{gopal2024separable}
& WACV'24
&0.894&0.637 & 0.866 & 0.669 & 0.448 & 0.335 & 0.744 & 0.531
& 0.858 & 0.680 & 0.457 & 0.323 & 0.620 & 0.460\\

AQATrack ~\cite{xie2024autoregressive}
& CVPR'24
&0.937&0.685 & 0.809 & 0.618 & 0.613 & 0.452 & 0.820 & 0.603

& 0.856 & 0.679 & 0.543 & 0.414 & \underline{0.720} & 0.544 \\

EVPTrack ~\cite{shi2024explicit}
& AAAI'24
&0.932&0.676 & 0.843 & 0.644 & 0.574 & 0.424 & 0.801 & 0.590
 & 0.876 & 0.701 & 0.534 & 0.413 & 0.710 & 0.535\\

ARTrackV2 ~\cite{bai2024artrackv2}
& CVPR'24
&0.912&0.622 & 0.910 & 0.667 & 0.524 & 0.375 & 0.798 & 0.547
& 0.868 & 0.675 & 0.536 & 0.414 & 0.687 & 0.522\\

\hline
MHT ~\cite{xiong2020material}
& TIP'20
&0.880&0.593
& 0.756 & 0.418 & 0.442 & 0.316 & 0.765 & 0.507
& 0.731 & 0.405 & 0.383 & 0.238 & 0.564 & 0.384 \\

TSCFW ~\cite{hou2022spatial}
& TGRS'22
&0.886&0.610
& 0.722 & 0.437 & 0.395 & 0.302 & 0.770 & 0.536
& 0.695 & 0.443 & 0.325 & 0.217 & 0.431 & 0.290 \\

BABS ~\cite{islam2023background}
& GRSL'23
&0.899&0.622
& 0.828 & 0.513 & 0.427 & 0.315 & 0.508 & 0.347
& 0.829 & 0.518 & 0.455 & 0.321 & 0.513 & 0.352\\

\hline

SPIRIT ~\cite{chen2023spirit}
& TGRS'23
&0.924&0.681
& 0.838 & 0.623 & 0.504 & 0.381 & 0.822 & 0.608
& 0.824 & 0.656 & 0.545 & 0.379 & 0.421 & 0.323 \\

SENSE ~\cite{chen2024sense}
& IF'24
&0.950&0.690 & 0.771 & 0.546 & 0.513 & 0.394 & 0.831 & 0.608
& 0.766 & 0.562 & 0.496 & 0.367 & 0.410 & 0.301\\

PHTrack ~\cite{chen2024phtrack}
& TGRS'24
&0.918& 0.663 & 0.780 & 0.535 & 0.528 & 0.415 & 0.800 & 0.581
& 0.732 & 0.528 & 0.395 & 0.264 & 0.423 & 0.306\\

MMF ~\cite{li2024material}
& TGRS'24
&0.931&0.692 & 0.910 & 0.666 & 0.410 & 0.324 & 0.815 & 0.613
& 0.875 & 0.701 & 0.521 & 0.395 & 0.645 & 0.488 \\

Trans-DAT ~\cite{wu2024domain}
& TCSVT'24
&--&-- & 0.900 & 0.683 & 0.726 & 0.535 & 0.725 & 0.511
 & 0.753 & 0.587 & 0.547 & 0.428 & 0.524 & 0.401\\

UBSTrack ~\cite{islam2025ubstrack}
& TGRS'25
&--&-- & -- & -- & -- & -- & -- & --  
& 0.861 & \underline{0.725} & 0.627 & 0.496 & 0.681 & 0.534 \\

ViPT ~\cite{zhu2023visual}
& CVPR'23
&0.909& 0.669& 0.950 & 0.742 & 0.709 & 0.563 & 0.858 & 0.641
& 0.851 & 0.691 & 0.650 & 0.503 & 0.716 & \underline{0.547} \\

HDSP ~\cite{HDSP}
& TGRS'25
&0.939 &0.688 & -- & -- & -- &-- & -- & -- 
& -- & -- & -- & -- & -- & --  \\

DaSSP-Net ~\cite{li2025multi}
& PR'25
&0.916&0.684 & \underline{0.952} & \underline{0.746} & \underline{0.731} & \underline{0.587} & 0.865 & 0.654 
& -- & -- & -- & -- & -- & --  \\

SSTtrack ~\cite{chen2025ssttrack}
& IF'25
&\underline{0.955}& \underline{0.713}& 0.854 & 0.660 & 0.512 & 0.400 & \underline{0.873} & \underline{0.657}
& \underline{0.889} & 0.711 & \underline{0.683} & 0.511 & 0.490 & 0.386\\
CSSTrack ~\cite{li2026CCStrack}
& PR'26
&0.951 &0.703 & -- & -- & -- &-- & -- & -- 
& -- & -- & -- & -- & -- & --  \\
\rowcolor{gray!15}
\textbf{E2E-MPT (Ours)}
& 
&\textbf{0.966} &\textbf{0.734} & \textbf{0.973} & \textbf{0.759} & \textbf{0.732} & \textbf{0.596} & \textbf{0.899} & \textbf{0.688}
& \textbf{0.939} & \textbf{0.758} & \textbf{0.724} & \textbf{0.549} & \textbf{0.743} & \textbf{0.564}\\
\hline
\end{tabular}}
\vspace{-8pt}
\end{table*}

\begin{figure*}[tbp]
    \centering
    \includegraphics[width=\linewidth]{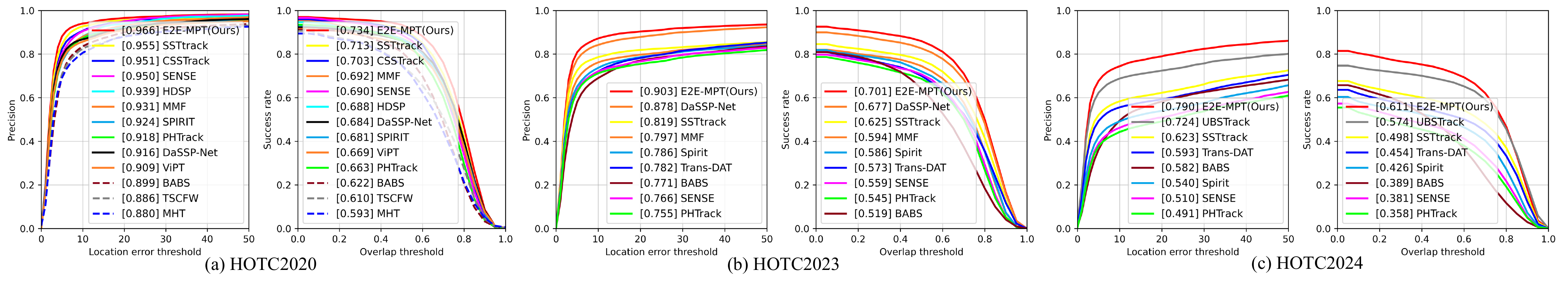}
    \vspace{-0.8cm}
    \caption{
    Performance comparison on three challenging hyperspectral object tracking benchmarks, including (a) HOTC2020, (b) HOTC2023, and (c) HOTC2024. 
    }
    \label{fig:hotcresults}
    \vspace{-10pt}
\end{figure*}

\noindent\textbf{Datasets.}
We evaluate on three standard 
benchmarks. HOTC2020~\cite{xiong2020material} contains 40 training and 35 testing sequences recorded at 25 FPS, with hyperspectral videos spanning 16 spectral channels (460--600nm) and annotated with 11 tracking attributes. HOTC2023 extends HOTC2020 across three hyperspectral 
domains: visible (VIS, 16 channels), near-infrared (NIR, 25 channels), and red-to-near-infrared (RedNIR, 15 channels),  with 110 training and 87 testing sequences. HOTC2024 further expands the benchmark to 216 training and 117 validation sequences under more challenging conditions. Details of these datasets can be found in~\cite{xiong2020material, li2025multi, islam2025ubstrack}.

\noindent\textbf{Metrics.}
We follow the one-pass evaluation (OPE) protocol~\cite{zhang2016robust} and report two standard metrics: the area under the curve (AUC) of the success plot and the distance precision (DP) at a 20-pixel threshold. AUC measures average overlap between predicted and ground-truth bounding boxes across IoU thresholds, while DP reports the percentage of frames whose center location error falls within 20 pixels.

\noindent\textbf{Baselines.}
We compare E2E-MPT against 6 RGB trackers: SeqTrack~\cite{chen2023seqtrack}, TCTrack++~\cite{cao2023towards}, SMAT~\cite{gopal2024separable}, AQATrack~\cite{xie2024autoregressive}, EVPTrack~\cite{shi2024explicit}, and ARTrackV2~\cite{bai2024artrackv2}; and 14 hyperspectral trackers: MHT~\cite{xiong2020material}, TSCFW~\cite{hou2022spatial}, BABS~\cite{islam2023background}, SPIRIT~\cite{chen2023spirit}, ViPT~\cite{zhu2023visual}, HDSP~\cite{HDSP}, MMF~\cite{li2024material}, Trans-DAT~\cite{wu2024domain}, PHTrack~\cite{chen2024phtrack}, SENSE~\cite{chen2024sense}, DaSSP-Net~\cite{li2025multi}, UBSTrack~\cite{islam2025ubstrack}, SSTtrack~\cite{chen2025ssttrack}, and CSSTrack~\cite{li2026CCStrack}. Among the hyperspectral methods, MHT~\cite{xiong2020material}, TSCFW~\cite{hou2022spatial}, and BABS~\cite{islam2023background} rely on hand-crafted features, while the remaining methods employ deep learning-based representations.

\noindent\textbf{Implementation details.}
E2E-MPT is implemented in PyTorch and trained on a single NVIDIA Tesla V100 GPU. We adopt the pretrained OSTrack backbone~\cite{ye2022joint} and follow the ViPT training protocol~\cite{zhu2023visual}, freezing the backbone parameters and optimizing only the newly introduced modules. The framework is trained end-to-end for 50 epochs with 5,600 template-search pairs sampled per epoch and a batch size of 16. We use AdamW optimizer~\cite{loshchilov2017decoupled} with an initial learning rate of $4\times10^{-5}$, decayed by a factor of 10 after 47 epochs, and a weight decay of $1\times10^{-4}$. Unless otherwise specified, the unmixing loss weight $\lambda_u$ and the target-region weighting factor $\lambda_{\mathrm{ce}}$ are set to 0.5 and 0.2, respectively.

\begin{figure}[tbp]
\vspace{-0.2cm}
    \centering
    \includegraphics[width=\columnwidth]{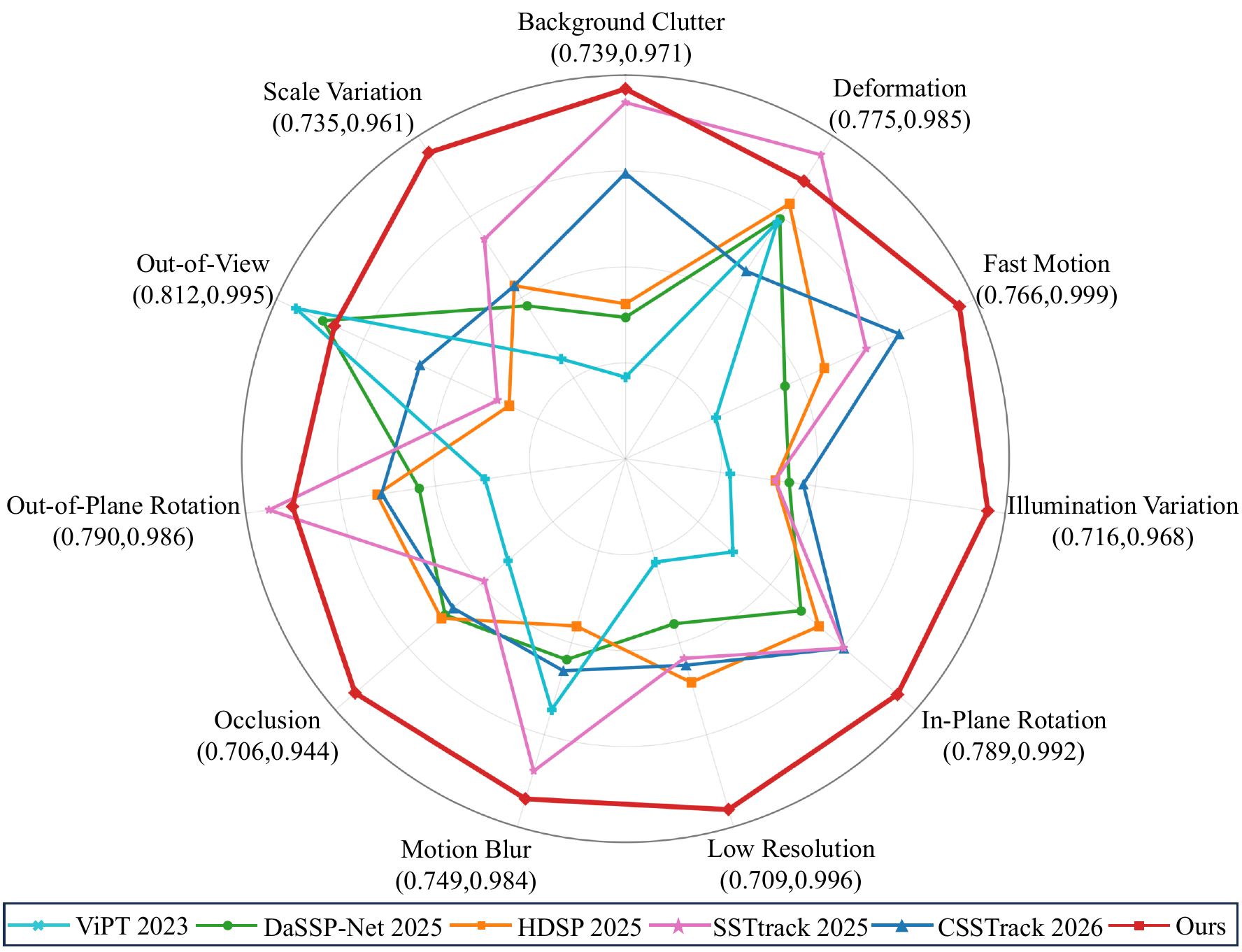}
    \caption{Attribute-based comparison on HOTC2020. 
The radar plot reports AUC scores of different methods, while the values in parentheses denote the AUC and DP scores of our method under each attribute as (AUC, DP).}
    \label{fig:attr_2020}
    \vspace{-0.3cm}
\end{figure}

\begin{figure*}[tbp]
    \centering
    \includegraphics[width=\textwidth]{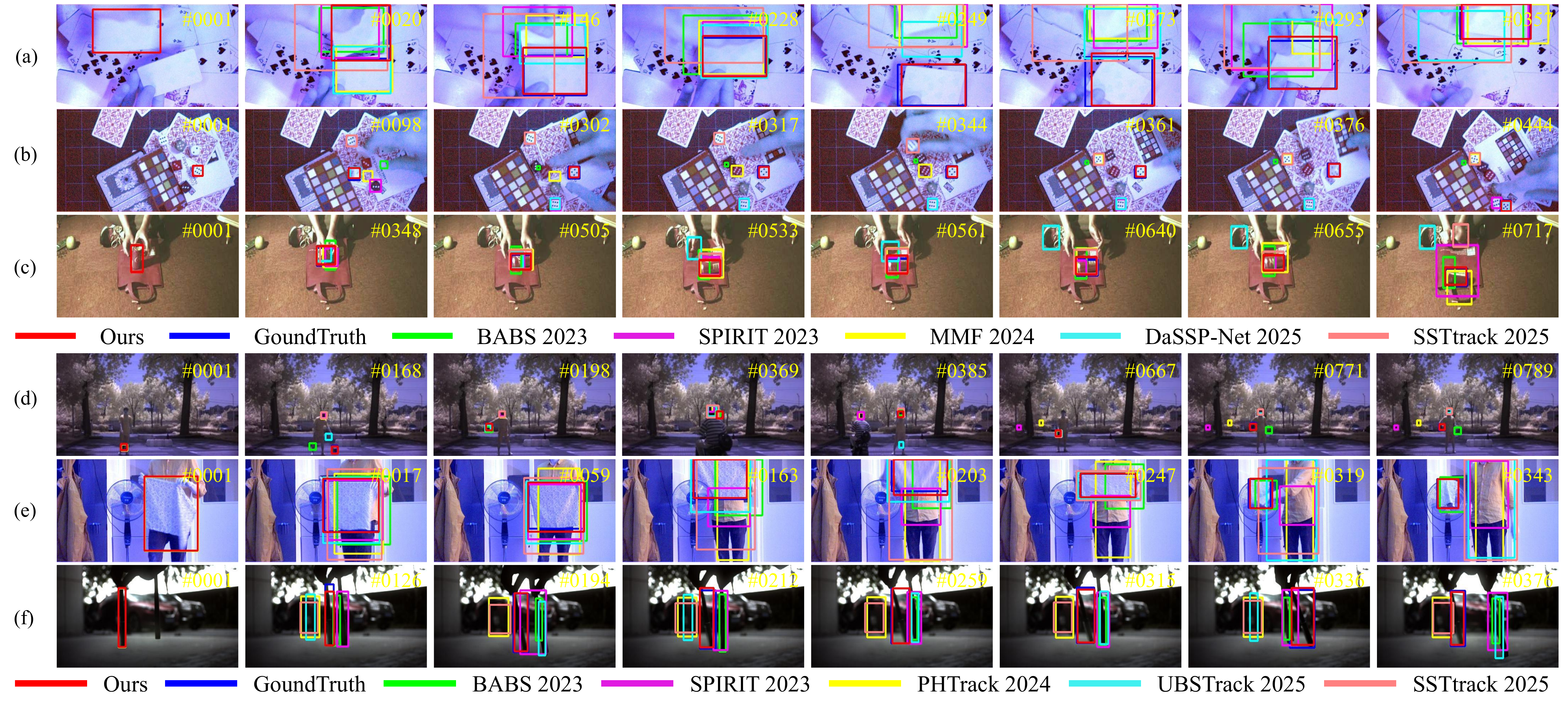}
    \vspace{-0.6cm}
    \caption{
Qualitative comparison of E2E-MPT against state-of-the-art trackers on six challenging sequences: (a) \textit{rednir-cards16}, (b) \textit{rednir-dice2}, (c) \textit{vis-coke}, (d) \textit{nir-basketball2}, (e) \textit{rednir-cloth3}, and (f) \textit{vis-pen4}. Sequences (a)--(c) are from HOTC2023, while (d)--(f) are from HOTC2024. Frame indices are shown in the top-left corner. For visualization, tracking results are overlaid on false-color images converted from hyperspectral inputs.
}

    \label{fig:qulitative}
\vspace{-10pt}
\end{figure*}

\subsection{Comparison with State-of-the-Art Methods}

\noindent\textbf{Comparison with RGB-based trackers.}
For fair comparison, false-color images (FCIs) serve as inputs for RGB-based methods when RGB images are unavailable. As shown in Table~\ref{tab:hotc_results}, E2E-MPT consistently outperforms all RGB trackers across all three benchmarks in both DP and AUC. On HOTC2020, E2E-MPT achieves 0.966/0.734 (DP/AUC), surpassing the strongest RGB competitor, SeqTrack~\cite{chen2023seqtrack}, by 2.0\%/4.5\%. On HOTC2023, 
E2E-MPT obtains 0.973/0.759, 0.732/0.596, and 0.899/0.688 on NIR, RedNIR, and VIS, respectively, exceeding all RGB baselines across all three spectral domains. On HOTC2024, E2E-MPT reaches 0.939/0.758, 0.724/0.549, and 0.743/0.564 on NIR, RedNIR, and VIS, respectively. These results confirm that directly exploiting hyperspectral material information yields stronger discriminative representations than FCI-based RGB tracking, which inherently suffers from spectral distortion due to band selection and channel compression.

\noindent\textbf{Comparison with hyperspectral trackers.}
As shown in Table~\ref{tab:hotc_results} and Fig.~\ref{fig:hotcresults}, E2E-MPT achieves state-of-the-art performance across all hyperspectral trackers on all three benchmarks. On HOTC2020, E2E-MPT obtains 0.966/0.734 (DP/AUC), surpassing SSTtrack~\cite{chen2025ssttrack} by 1.1\%/2.1\% and CSSTrack~\cite{li2026CCStrack} by 1.5\%/3.1\%, as reflected by the highest precision and success curves in Fig.~\ref{fig:hotcresults}(a). On HOTC2023, E2E-MPT achieves the best results across all three spectral domains, reaching 0.973/0.759, 0.732/0.596, and 0.899/0.688 on NIR, RedNIR, and VIS, respectively. Against the strongest per-domain competitors, E2E-MPT improves over DaSSP-Net~\cite{li2025multi} by 2.1\%/1.3\% on NIR and 0.1\%/0.9\% on RedNIR, and over SSTTrack~\cite{chen2025ssttrack} by 2.6\%/3.1\% on VIS. In terms of overall performance, E2E-MPT achieves 0.903/0.701, outperforming DaSSP-Net~\cite{li2025multi} by 2.5\%/2.4\%, as shown in Fig.~\ref{fig:hotcresults}(b). On HOTC2024, E2E-MPT again yields the best results across all three domains, obtaining 0.939/0.758, 0.724/0.549, and 0.743/0.564 on 
NIR, RedNIR, and VIS, respectively, with an overall score of 0.790/0.611. Compared with the strongest overall competitor UBSTrack~\cite{islam2025ubstrack}, E2E-MPT improves by 6.6\%/3.7\% in DP/AUC, as shown in Fig.~\ref{fig:hotcresults}(c). Notably, compared with MMF~\cite{li2024material}, the closest material-aware method that relies on a decoupled external unmixing pipeline, E2E-MPT improves performance by 6.3\%/9.3\% on HOTC2023 NIR, directly demonstrating the advantage of end-to-end joint optimization over decoupled material-aware tracking. These consistent gains across spectral domains and benchmark years confirm the superiority of the proposed framework in both localization accuracy and overlap-based robustness.

\begin{table}[tbp]
\centering
\vspace{-0.3cm}
\small
\setlength{\tabcolsep}{8pt}
\renewcommand{\arraystretch}{1.05}
\caption{Component-wise ablation of our framework on HOTC2020. 
}

\label{tab:component_ablation}
\begin{tabular}{c | cc}
\hline
Method & DP & AUC \\
\hline
Baseline 
& 0.910 & 0.662 \\

Baseline + MCP
& 0.915 & 0.692 \\

Baseline + DWMPM
& 0.927 &  0.702\\

Baseline + MRDM + MCP
&  0.941 & 0.713 \\
Baseline + MRDM + DWMPM
&  0.951 &  0.720\\
Baseline + MRDM + MCP + FPFM
&  0.961 &  0.727\\
\rowcolor{gray!15}
Baseline + MRDM + DWMPM + FPFM
& \textbf{0.966} & \textbf{0.734} \\
\hline
\end{tabular}

\vspace{-0.2cm}
\end{table}

\begin{table}[tbp]
\centering

\vspace{-0.2cm}

\small
\setlength{\tabcolsep}{8pt}
\renewcommand{\arraystretch}{1.05}
\caption{Ablation study on different unmixing strategies evaluated on HOTC2020 and HOTC2023 datasets.}

\label{tab:unmixing_ablation}
\begin{tabular}{l | cc | cc}
\hline
\multirow{2}{*}{Method}
& \multicolumn{2}{c|}{HOTC2020} 
& \multicolumn{2}{c}{HOTC2023} \\
& DP & AUC & DP & AUC \\
\hline
CLSUnSAL+MMF
& 0.933 & 0.691 
& 0.797 & 0.595 \\
\textbf{CLSUnSAL+MPT}

& \textbf{0.939} & \textbf{0.701} 
& \textbf{0.839} & \textbf{0.642} \\
\hline
DAEU+MPT
& 0.934 &  0.709
& 0.884 & 0.686 \\
\textbf{DAEU+E2E-MPT }
& \textbf{0.943} & \textbf{0.718} 
& \textbf{0.894} & \textbf{0.695} \\

\hline
CNNAEU+MPT
& 0.942 &  0.711
& 0.897  & 0.697 \\
\rowcolor{gray!15}
\textbf{CNNAEU+E2E-MPT}

& \textbf{0.966} & \textbf{0.734}
& \textbf{0.902} & \textbf{0.701} \\

\hline
\end{tabular}
\vspace{-0.5cm}
\end{table}

\noindent\textbf{Attribute analysis.}
Fig.~\ref{fig:attr_2020} reports attribute-based AUC scores under 11 challenging conditions on HOTC2020. E2E-MPT achieves the best performance on 8 of 11 attributes, including Background Clutter (BC), Fast Motion (FM), Illumination Variation (IV), In-Plane Rotation (IPR), Low Resolution (LR), Motion Blur (MB), Occlusion (OCC), and Scale Variation (SV). The most notable gains are observed under IV and LR, where material-level spectral cues remain stable despite appearance changes caused by lighting shifts and cluttered backgrounds, conditions under which RGB and FCI-based methods suffer most. Under FM and MB, the frequency-decomposed material prompts provide complementary structural cues that compensate for degraded spatial appearance. On the remaining three attributes, Deformation (DEF),  Out-of-Plane Rotation (OPR), and Out-of-View (OV), E2E-MPT remains competitive 
but does not achieve the top rank. This is expected, as these attributes primarily involve geometric transformations and target disappearance, which are less directly addressed by material-level spectral decomposition and remain an avenue for future work.

\noindent\textbf{Qualitative analysis.}
Fig.~\ref{fig:qulitative} presents qualitative comparisons on six representative sequences spanning the RedNIR, VIS, and NIR domains of HOTC2023 and HOTC2024. In sequences (a) \textit{rednir-cards16} and (b) \textit{rednir-dice2}, competing methods frequently drift toward cluttered background regions due to the low inter-class spectral contrast in the RedNIR domain, whereas E2E-MPT maintains accurate localization by leveraging target-oriented material decomposition to suppress background interference. In sequence (c) \textit{vis-coke}, where significant illumination variation degrades appearance-based cues, E2E-MPT produces stable bounding boxes while BABS~\cite{islam2023background}, SPIRIT~\cite{chen2023spirit}, and MMF~\cite{li2024material} exhibit notable drift. In sequences (d) \textit{nir-basketball2} and (e) \textit{rednir-cloth3} from HOTC2024, which involve fast motion and scale variation, E2E-MPT consistently localizes the target with tighter bounding boxes compared to PHTrack~\cite{chen2024phtrack} and UBSTrack~\cite{islam2025ubstrack}. In sequence (f) \textit{vis-pen4}, captured under low-light conditions, competing methods lose the target entirely in several frames, while E2E-MPT maintains robust tracking by exploiting spectral material cues that remain discriminative even under severe illumination degradation. These qualitative results collectively demonstrate that end-to-end material-aware prompting provides more stable and discriminative tracking responses across diverse hyperspectral domains and challenging scenarios.

\begin{table}[tbp]
\centering
\small
\vspace{-0.3cm}
\setlength{\tabcolsep}{8pt}
\renewcommand{\arraystretch}{1.05}
\caption{Ablation study of AMD and its components on HOTC2020. 
}

\label{tab:amd_ablation}
\begin{tabular}{c | c c}
\hline
Configuration & DP & AUC \\
\hline
Split channels (random) & 0.927 & 0.703 \\
1D Haar & 0.949 & 0.717 \\
Adaptor + Split channels & 0.937 & 0.713 \\
Adaptor + FFT & 0.950 & 0.716 \\
\rowcolor{gray!15}
\textbf{E2E-MPT} (Adaptor + 1D Haar) & \textbf{0.966} & \textbf{0.734} \\
\hline
\end{tabular}
\vspace{-0.5cm}
\end{table}

\begin{table}[tbp]
\centering
\vspace{-0.3cm}
\small
\setlength{\tabcolsep}{4.5pt}
\caption{Effect of prompt injection depth on computational complexity and tracking performance evaluated on HOTC2020.}
\label{tab:prompt_layers}
\begin{tabular}{c | c | c | c | c c}
\hline
\#Layers & MACs (G) & Params (M) & FPS & DP & AUC \\
\hline
1  & 23.51 & 0.78 & 48.14 &  0.923& 0.698 \\
1-3  &  23.59 & 1.11 & 36.27 & 0.935 & 0.709 \\
1-6  & 23.72 & 1.62  & 28.93 & 0.942 & 0.716 \\
1-9  & 23.84 &  2.12 & 22.84 & 0.950 &  0.718\\
1,6,12 & 23.59 & 1.11 & 36.27  &   0.916    &   0.694  \\
1,3,6,9,12 &23.67&1.45 &29.82 &   0.953   &   0.723    \\
\rowcolor{gray!15}
1,2,4,6,8,10,12 &23.76 &1.78 &26.02 &    \underline{0.966}  &  \textbf{0.734}  \\
\rowcolor{gray!15}
1-12 & 23.96 & 2.63 & 20.73 & \textbf{0.967} & \underline{0.730} \\
\hline
\end{tabular}
\end{table}

\subsection{Ablation Study}
\noindent\textbf{Effectiveness of key components.}
Table~\ref{tab:component_ablation} reports the contribution of each proposed component. The baseline fine-tunes only a lightweight projection block to map hyperspectral inputs into the backbone embedding space with the backbone frozen. When MRDM is excluded, false-color images serve as the backbone input and projected hyperspectral features are used directly for prompt generation. As shown, each component brings consistent improvements: DWMPM outperforms the MCP prompt~\cite{zhu2023visual} under both settings, MRDM provides additional gains by supplying tracking-oriented material representations, and FPFM further improves performance through adaptive frequency-branch fusion. The full model achieves the best results across all metrics, confirming that the three components contribute complementary and non-redundant benefits.

\begin{table}[tbp]
\vspace{-0.3cm}
\centering
\caption{Ablation study of different operator assignments for low- and high-frequency branches in the DWMPM on HOTC2020. The row corresponding to our design is shaded in gray.}
\label{tab:wmpm_ablation}

\renewcommand{\arraystretch}{1.05}
\setlength{\tabcolsep}{0.15em}
\resizebox{\linewidth}{!}{
\begin{tabular}{ c c | c c | c c}
\hline
Low Frequency & High Frequency & MACs (G) & Params (M) & DP & AUC \\
\hline
Conv       & Conv       & 23.76 & 1.79 & 0.952 & 0.720 \\
Cross-Attn & Cross-Attn & 23.77 & 1.85 & 0.948 & 0.719 \\
Conv       & Cross-Attn & 23.77 & 1.77 & 0.957 & 0.723 \\
\rowcolor{gray!15}
Cross-Attn & Conv       & 23.76 & 1.78 & \textbf{0.966} & \textbf{0.734} \\
\hline
\end{tabular}}
\vspace{-0.3cm}
\end{table}

\noindent\textbf{End-to-end unmixing and prompting.}
Table~\ref{tab:unmixing_ablation} evaluates the MPT design and end-to-end joint optimization across different unmixing backbones. Using the same material cues from CLSUnSAL~\cite{iordache2013collaborative}, replacing MMF~\cite{li2024material} with our MPT consistently improves both DP and AUC on HOTC2020 and HOTC2023, confirming that the proposed prompt design more effectively exploits material information for tracking. For both DAEU~\cite{palsson2018hyperspectral} and CNNAEU~\cite{palsson2020convolutional}, end-to-end joint optimization consistently outperforms the two-stage counterpart on both datasets, demonstrating that jointly optimizing unmixing and tracking aligns material representations with the localization objective. The full E2E-MPT with CNNAEU~\cite{palsson2020convolutional} achieves the best overall performance, reaching 0.966/0.734 on HOTC2020 and 0.902/0.701 on HOTC2023. These results confirm that E2E-MPT is both effective and generalizable across different hyperspectral unmixing backbones.

\noindent\textbf{Impact of AMD.}
Table~\ref{tab:amd_ablation} compares different abundance map decomposition strategies on HOTC2020. Random channel splitting yields the weakest performance (0.927/0.703), confirming that naive partitioning fails to decouple complementary material cues effectively. Replacing random splitting with 1D Haar wavelet transform improves performance to 0.949/0.717, demonstrating that frequency-aware decomposition provides more structured material representations. Introducing the channel adaptor before splitting brings further gains (0.937/0.713), indicating that re-projecting abundance channels into a tracking-conditioned basis better aligns material representations with the backbone feature space. Among frequency decomposition strategies, 1D Haar outperforms FFT (0.966/0.734 vs. 0.950/0.716), as 1D Haar preserves local continuity and adjacent variations along the abundance-channel dimension, while FFT represents the channel sequence using global frequency bases and is less sensitive to localized material transitions. The full AMD configuration, combining the adaptor and 1D Haar, achieves the best performance, confirming that channel adaptation and frequency decomposition provide complementary and mutually reinforcing benefits.

\begin{table}[tbp]
\centering


\small
\setlength{\tabcolsep}{3.5pt}
\renewcommand{\arraystretch}{1.15}
\caption{Ablation study on the effectiveness of FPFM on HOTC2020 }
\label{tab:fpfm_ablation}
\begin{tabular}{c | c | c |c|c c}
\hline
Fusion Strategy & MACs (G) & Params (M)  & DP & AUC \\
\hline
   
  Addition   & 23.7 &1.7 & 0.951 & 0.720 \\ 
   Convolution Fusion &24.1  &2.9  & 0.952 & 0.701 \\
   Attention Fusion &24.0  & 2.6 & 0.945 & 0.716\\
   \rowcolor{gray!15}
    FPFM& 24.0  &2.6 & \textbf{0.966} & \textbf{0.734} \\

\hline
\end{tabular}

\end{table}

\begin{figure}[tbp]
    \centering
    \vspace{-0.3cm}
    \includegraphics[width=\linewidth]{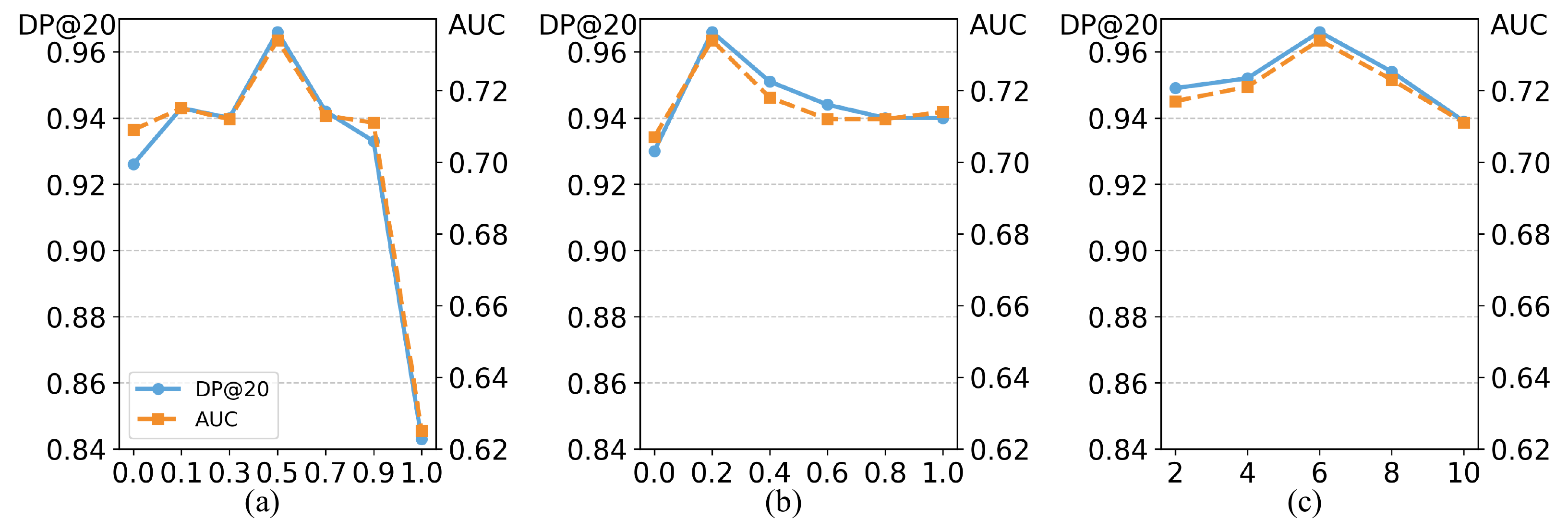}
    \vspace{-0.3cm}
    \caption{
    Ablation study of loss parameters on the HOTC2020 dataset.
    (a) Impact of the unmixing loss weight $\lambda_u$.
    (b) Effect of target-oriented loss weight $\lambda_{\mathrm{ce}}$.
    (c) Influence of different endmember numbers $r$.
    }
    \label{fig:loss_ablation}
    \vspace{-0.3cm}
\end{figure}

\begin{figure}[tbp]
    \centering
    
    \includegraphics[width=\linewidth]{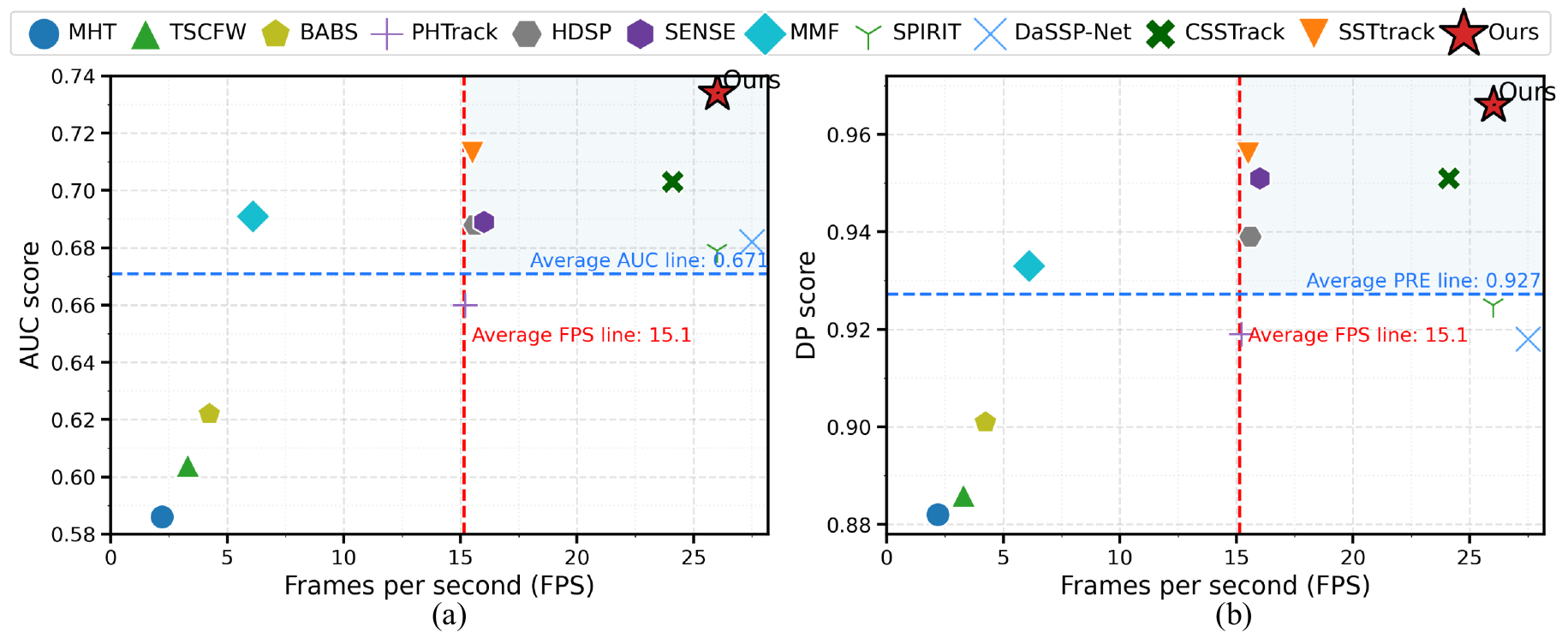}
    \vspace{-0.4cm}
    \caption{
   Tracking efficiency comparison on HOTC2020.
}
    \label{fig:complexity}
\vspace{-0.4cm}
\end{figure}

\begin{table}[tbp]
\centering
\small
\setlength{\tabcolsep}{3.5pt}
\caption{Computational complexity and tracking performance comparison. OSTrack is the foundation model.}
\label{tab:complexity}
\begin{tabular}{c c c c |c c}
\hline
Method  & MACs (G)   & FPS &  Params (M) & DP & AUC \\
\hline
OSTrack   & 21.5    & 52.4 & 93.4 & 0.910 & 0.662 \\
\hline
ViPT      & 21.8    & 33.8 & +0.8 & 0.909 & 0.669 \\
DaSSPNet  & 24.7    & 27.8 & +2.2 & \underline{0.917} & \underline{0.682} \\
\rowcolor{gray!15}
\textbf{E2E-MPT (Ours)}& 24.0 & 26.0  & +1.8 & \textbf{0.966} & \textbf{0.734} \\
\hline
\end{tabular}
\vspace{-0.3cm}
\end{table}

\noindent\textbf{Impact of injection layers 
}
Table~\ref{tab:prompt_layers} reports the effect of injecting DWMPM at different layers. Introducing DWMPM into more layers generally improves performance, though the injection pattern also matters. Applying DWMPM to all layers achieves the highest DP (0.967) but at the cost of significantly reduced inference speed (20.73 FPS). The selected configuration \{1,2,4,6,8,10,12\} achieves a slightly higher AUC (0.734) with nearly identical DP (0.966) while running at 26.02 FPS, offering a better trade-off between accuracy and efficiency. This configuration is adopted as the default setting. 

\noindent\textbf{Operator design in WMP blocks.}
Table~\ref{tab:wmpm_ablation} validates the asymmetric operator design in the WMP block by comparing four configurations: convolution for both branches, cross-attention for both branches, swapped operators, and the proposed design (cross-attention for low-frequency, convolution for high-frequency). Computational cost and parameter size are kept comparable across all variants. The proposed design achieves the best performance, confirming that cross-attention is better suited for modeling globally distributed low-frequency material cues, while convolution more effectively captures spatially local high-frequency variations. This is consistent with the design rationale discussed in Section~\ref{subsec:wmp}.

\noindent\textbf{Impact of FPFM.}
Table~\ref{tab:fpfm_ablation} compares FPFM against alternative fusion strategies: element-wise addition, multiplication, convolutional fusion, and attention fusion. Addition and 
multiplication introduce no extra parameters, while convolutional fusion concatenates the two branch prompts and processes them through three convolutional layers before residual injection. FPFM achieves the best performance 
at comparable computational cost, demonstrating that adaptive latent-space fusion more effectively balances complementary low- and high-frequency material cues than hand-crafted alternatives.

\noindent\textbf{Hyperparameter analysis.}
Fig.~\ref{fig:loss_ablation}(a) shows the effect of the unmixing loss weight $\lambda_u$. The best performance is achieved at $\lambda_u=0.5$, confirming that balanced joint optimization between tracking and unmixing objectives is critical. Setting $\lambda_u=1.0$ eliminates tracking supervision and yields the worst performance, while $\lambda_u=0$ degenerates the unmixing branch into a generic prompt generator without physically meaningful material representations, both resulting in inferior tracking.
Fig.~\ref{fig:loss_ablation}(b) shows the sensitivity of the target-oriented loss weight $\lambda_{\mathrm{ce}}$. Performance peaks at $\lambda_{\mathrm{ce}}=0.2$, where the balance between target-relevant and background regions in the search area is most effective; both smaller and larger values degrade performance.
Fig.~\ref{fig:loss_ablation}(c) shows the effect of the number of endmembers $r$, varied from 2 to 10. Performance improves with increasing $r$, peaks at $r=6$, and declines thereafter as over-parameterization introduces redundant material components. We therefore set $r=6$ in all experiments.

\begin{figure*}[tbp]
    \centering
    \includegraphics[width=\textwidth]{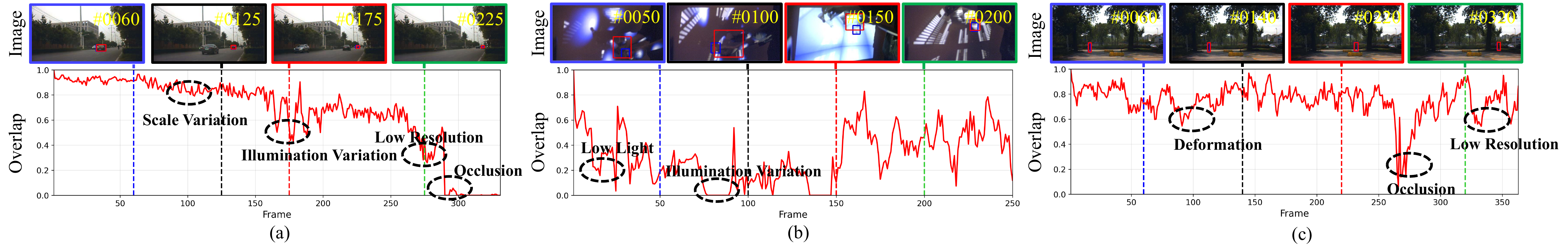}
    \vspace{-0.6cm}
    \caption{
    Overlap curves and representative tracking samples of E2E-MPT on HOTC2023. 
    (a) \textit{car3}, with attributes including scale variation (SV), illumination variation (IV), low resolution (LR), and occlusion (OCC). 
    (b) \textit{partylights6}, with low light (LL) and illumination variation (IV). 
    (c) \textit{pedestrian2}, with occlusion (OCC), low resolution (LR), and deformation (DEF).  
    The \textcolor{blue}{blue} and \textcolor{red}{red} bounding boxes denote the ground truth and the prediction, respectively.
    }
    \label{fig:failurecase}
    \vspace{-0.3cm}
\end{figure*}

\begin{figure}[tbp]
    \centering
    \includegraphics[width=\columnwidth]{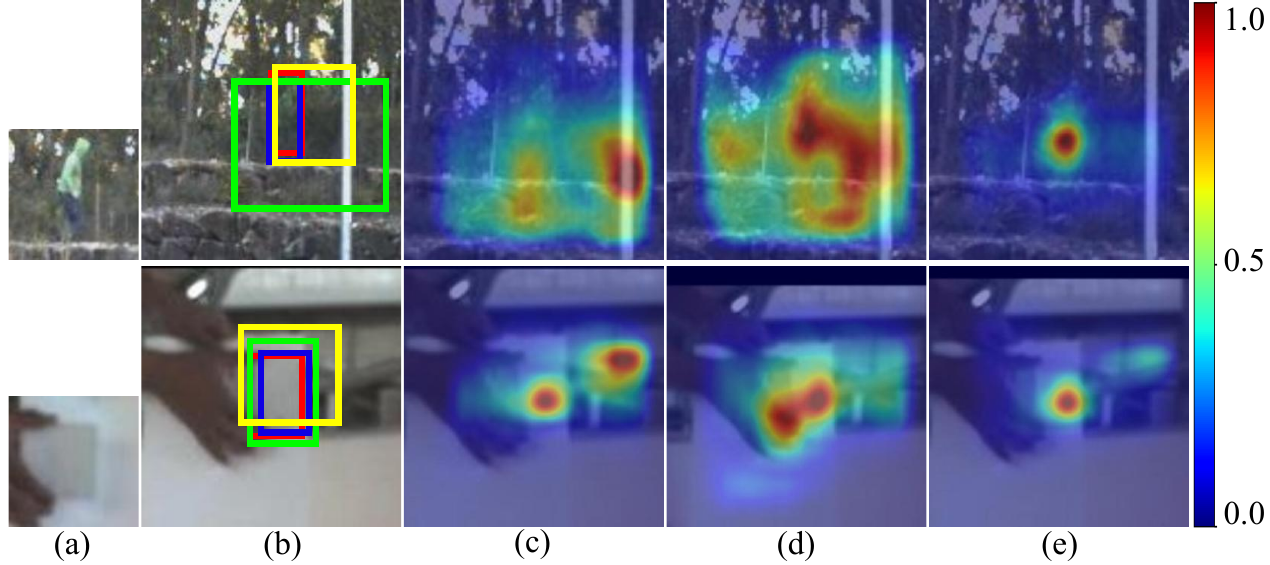}
    \vspace{-0.6cm}
    \caption{
    Qualitative visualization of tracking results and score maps.
    (a) Template visualization.
    (b) Tracking results in the search region, where the \textcolor{blue}{blue}, \textcolor{red}{red}, \textcolor{yellow}{yellow}, and \textcolor{green}{green} boxes denote the ground truth, the proposed end-to-end method, the non-end-to-end variant, and the baseline, respectively.
    (c) Score map of OSTrack (baseline).
    (d) Score map of the non-end-to-end variant (CNNAEU + MPT).
    (e) Score map of the proposed end-to-end method.
    }
    \vspace{-0.3cm}
    \label{fig:visual_results}
\end{figure}

\subsection{Analysis \& Discussion}

\noindent\textbf{Complexity and efficiency.}
Table~\ref{tab:complexity} summarizes the computational cost and tracking performance of compared methods. Relative to the foundation model OSTrack, E2E-MPT improves AUC from $0.662$ to $0.734$ and DP from $0.910$ to $0.966$ with only $1.78$M additional parameters and moderate computational overhead. Among prompt-based hyperspectral trackers, E2E-MPT outperforms both ViPT~\cite{zhu2023visual} and DaSSP-Net~\cite{li2025multi} in AUC and DP at comparable complexity, indicating that the performance gains stem from the proposed material-aware prompting and end-to-end optimization rather than model scale.
Fig.~\ref{fig:complexity} further illustrates the accuracy-efficiency trade-off on HOTC2020. Several competing methods fall below the average FPS threshold of 15.1 or fail to exceed the average AUC of 0.671 and DP of 0.927. E2E-MPT achieves the highest AUC and DP scores while maintaining competitive inference speed, demonstrating a favorable trade-off between tracking accuracy and computational efficiency.

\noindent\textbf{Failure case analysis.}
Fig.~\ref{fig:failurecase} shows representative failure cases on HOTC2023. Under long-term occlusion and low-resolution conditions (Figs.~\ref{fig:failurecase}(a) and (c)), tracking performance degrades significantly, as the material-driven framework lacks explicit temporal modeling to recover the target when appearance is severely degraded or invisible over extended periods. Under extreme illumination changes (Fig.~\ref{fig:failurecase}(b)), single-frame material cues become unstable and insufficiently discriminative for reliable localization. These limitations suggest that incorporating temporal cues such as motion continuity or tracking memory represents a promising direction for future work.

\noindent\textbf{Score map.}
Fig.~\ref{fig:visual_results} visualizes tracking score maps under challenging scenarios. Compared with OSTrack and the non-end-to-endvariant, E2E-MPT produces more concentrated and less ambiguous response maps around the target region, demonstrating that the target-oriented unmixing loss improves foreground-background separation and suppresses background-dominated responses. These visualizations provide qualitative evidence that end-to-end joint optimization directly benefits target localization beyond what quantitative metrics alone convey.

\noindent\textbf{Limitation.}
E2E-MPT has two primary limitations. First, the unmixing network may fail to reliably decompose materials that are absent or underrepresented in the training distribution, which degrades the quality of material cues and consequently affects tracking performance. Second, the framework does not exploit temporal information, making it vulnerable under long-term occlusion and extreme illumination conditions where target material cues cannot be reliably inferred from a single frame. Incorporating temporal modeling is an important direction for future work.
\section{Conclusion}
\label{sec:conclusion}
We presented E2E-MPT, an end-to-end framework for hyperspectral object tracking that jointly optimizes material decomposition and localization. The results show that material cues can better support tracking when material decomposition is aligned with the localization objective and integrated into the backbone through structured prompt design, rather than being extracted by a decoupled external unmixing pipeline. At the same time, the framework still depends on the quality of unmixing and may degrade when unseen materials are not well represented by the learned decomposition model. In addition, the current design does not explicitly exploit temporal information, which limits its robustness under long-term occlusion and severe illumination variation. Future work will focus on incorporating temporal modeling to further improve robustness in these challenging scenarios.

\bibliographystyle{IEEEtran}
\bibliography{reference}

\vskip -2.5\baselineskip plus -1fil

\begin{IEEEbiography}[{\includegraphics[trim=70 350 70 0,width=1in,height=1.25in,clip,keepaspectratio]{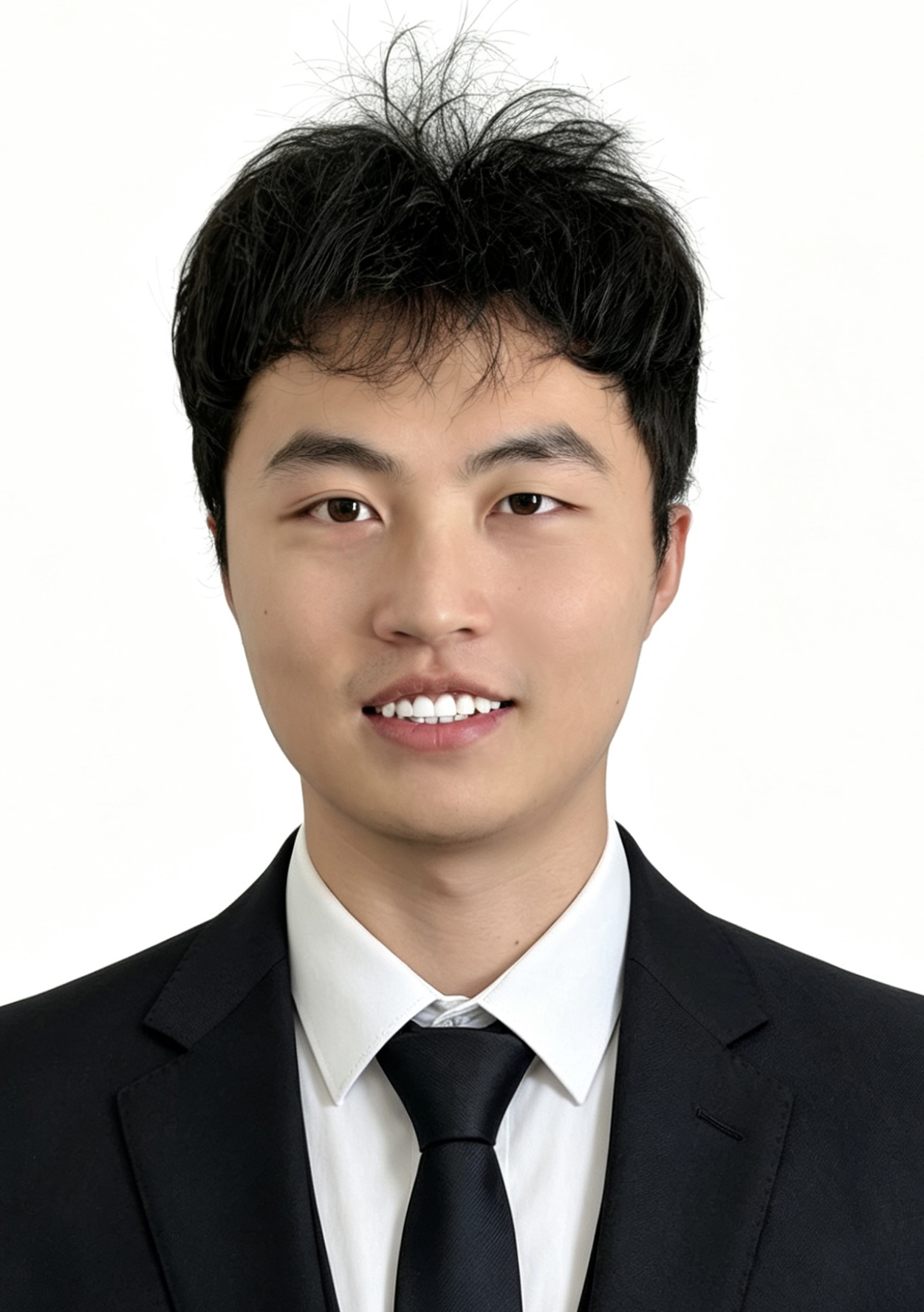}}]{Xu Han}

received the Master of  degree in Information Technology from the University of New South Wales, Sydney, NSW, Australia, in 2024. He is currently pursuing the Ph.D. degree with the School of Information and Communication Technology at Griffith University. His research interests include hyperspectral object tracking, data compression and continual learning. 

\end{IEEEbiography}

\vskip -2.5\baselineskip plus -1fil

\begin{IEEEbiography}[{\includegraphics[trim=30 20 30 35,width=1in,height=1.25in,clip,keepaspectratio]{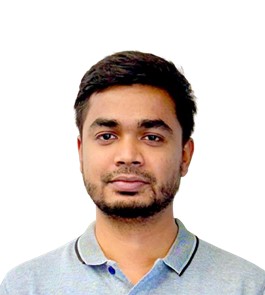}}]{Mohammad Aminul Islam} received the B.Sc. (Engg.) degree in computer science and engineering from University of Chittagong, Chattogram, Bangladesh, in 2013, and the M.S. degree in computer science from Bangladesh Agricultural University, Mymensingh, Bangladesh, in 2019. He received the Ph.D. degree from Griffith University, Australia, in 2025.
In March 2026, he joined the School of Information and Communication Technology at Griffith University, Nathan, QLD, Australia, where he is currently a Research Fellow. He is also an Assistant Professor with the Department of Computer Science and Mathematics, Bangladesh Agricultural University, Mymensingh, Bangladesh.
His research interests include spectral imaging, hyperspectral object tracking, information fusion, computer vision, and pattern recognition.
\end{IEEEbiography}

\vskip -2.5\baselineskip plus -1fil

\begin{IEEEbiography}
[{\includegraphics[width=1in,height=1.25in,clip,keepaspectratio]{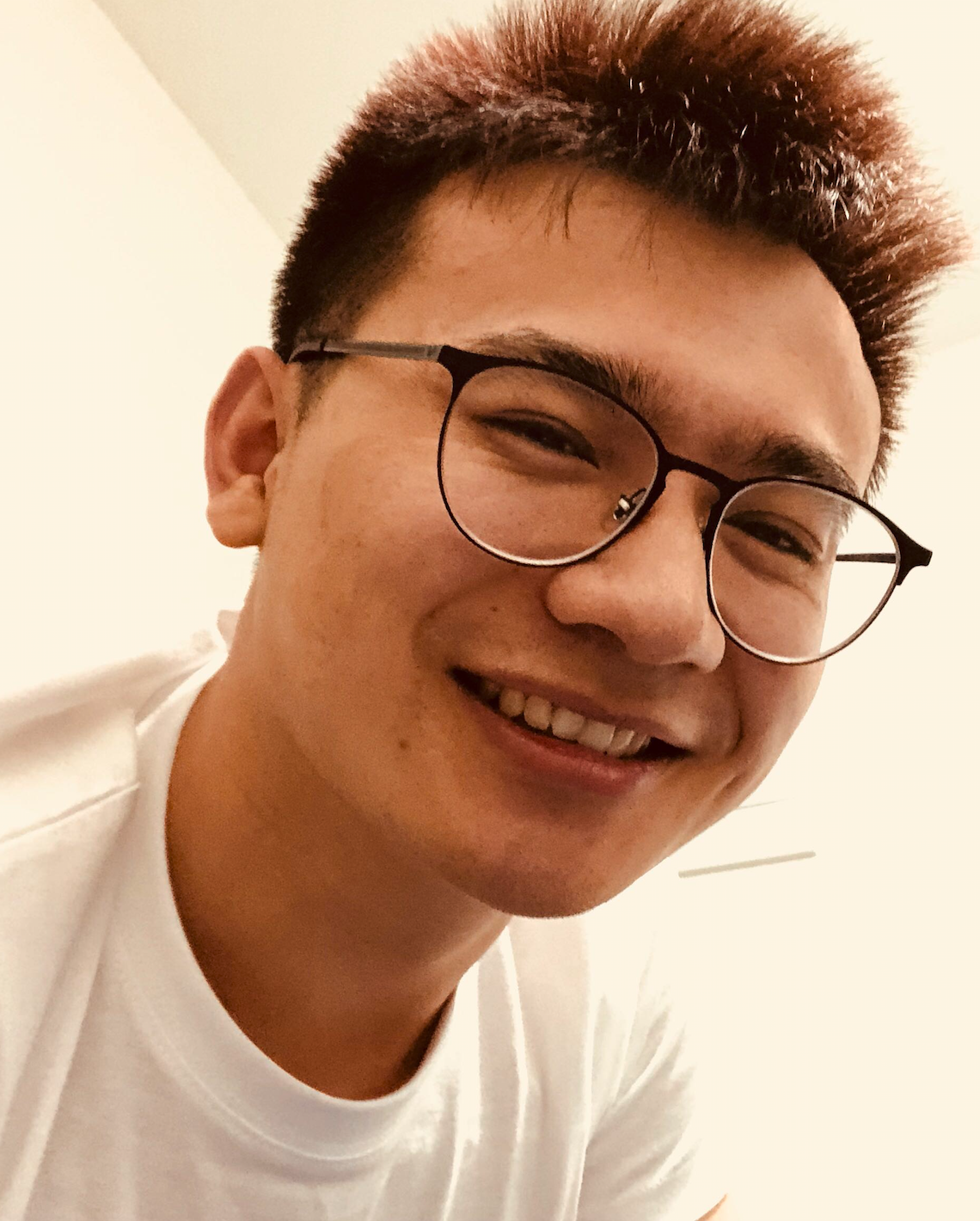}}]{Lei Wang} received his M.E. in Software Engineering from the University of Western Australia (UWA) in 2018 and his Ph.D. in Engineering and Computer Science from the Australian National University (ANU) in 2023. He is a Research Fellow in the School of Electrical and Electronic Engineering at Griffith University and a Visiting Scientist with Data61/CSIRO. He leads the Temporal Intelligence and Motion Extraction (TIME) Lab at Griffith University. He previously held research positions at ANU, UWA, and Data61/CSIRO. His research focuses on motion-, data-, and model-centric approaches to action recognition and anomaly detection. 
He has authored numerous first-author papers in top-tier venues, including CVPR, ICCV, ECCV, ACM Multimedia, NeurIPS, ICLR, ICML, AAAI, TPAMI, IJCV, and TIP, and received the Sang Uk Lee Best Student Paper Award at ACCV 2022. He serves as an Area Chair for NeurIPS 2026, ACM Multimedia 2024-2026, ICASSP 2025, and ICPR 2024, and was recognized as an Outstanding Area Chair at ACM Multimedia 2024.
\end{IEEEbiography}

\vskip -2.5\baselineskip plus -1fil

\begin{IEEEbiography}[{\includegraphics[trim=2 12 35 12,width=1.3in,height=1.6in,clip,keepaspectratio,angle=-90]{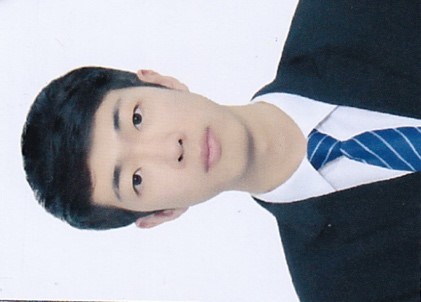}}]{Zekun Long} received the Master of Information Technology degree in Data Analytics from Griffith University, Brisbane, QLD, Australia, in 2022. He is currently pursuing the Ph.D. degree with the School of Information and Communication Technology at Griffith University. His research interests include hyperspectral image processing, spectral unmixing, and remote sensing analytics. His recent work focuses on weak-signal hyperspectral modeling, deep learning architectures, and diffusion-based spectral priors for quantitative environmental and food-safety analysis.

\end{IEEEbiography}

\vskip -2.5\baselineskip plus -1fil

\begin{IEEEbiography}[{\includegraphics[trim=120 280 130 60,width=1in,height=1.25in,clip,keepaspectratio]{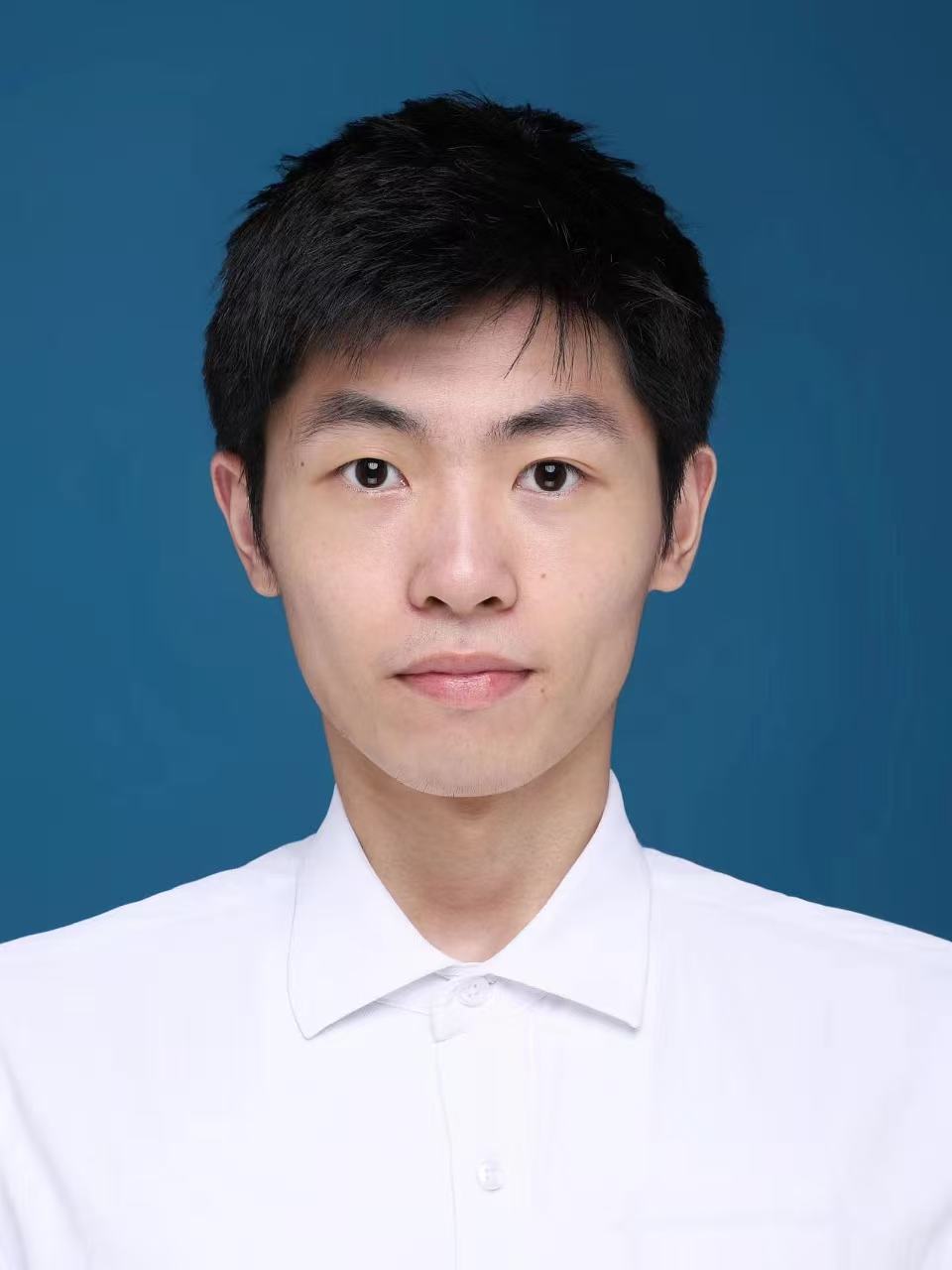}}]{Guanmanyi Fu}
received the B.E. degree in bioengineering from Northwest A\&F University, Xianyang, China, in 2017, and the Ph.D. degree in computer science and technology from the School of Computer Science and Engineering, Nanjing University of Science and Technology, Nanjing, China, in 2026. He is currently a Research Fellow at Griffith University, QLD, Australia. His research interests include image processing and machine learning.
\end{IEEEbiography}

\vskip -2.5\baselineskip plus -1fil

\begin{IEEEbiography}[{\includegraphics[trim=5 30 6 5,width=1in,height=1.25in,clip,keepaspectratio]{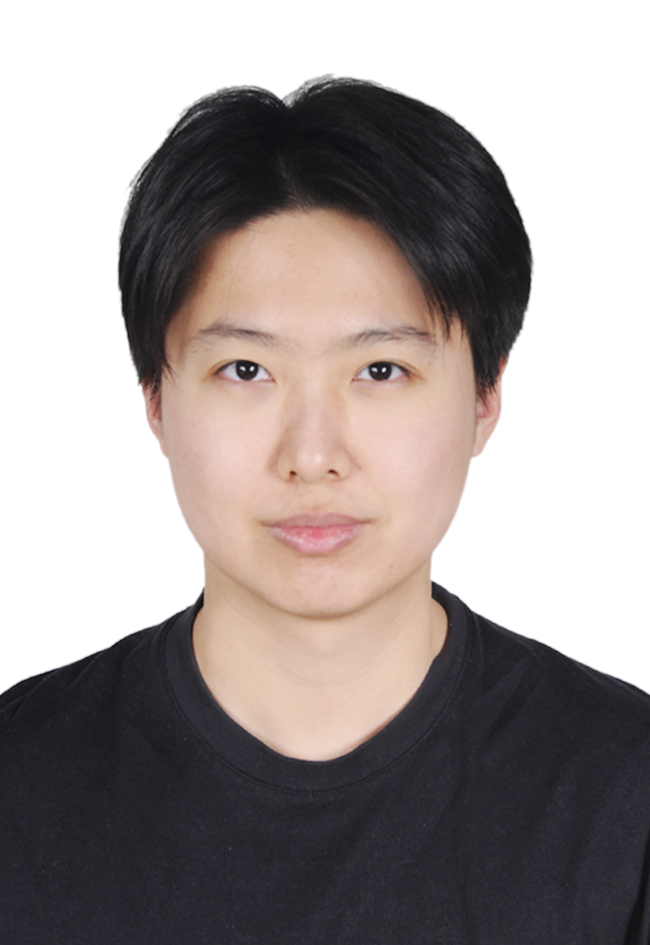}}]{Wangshu Cai} received the Master of Machine Learning and Computer Vision degree from the Australian National University, Canberra, ACT, Australia in 2024. He is currently pursuing the Ph.D. degree with the School of Information and Communication Technology as Griffith University. His research interests include hyperspectral object tracking and generative AI.

\end{IEEEbiography}

\vskip -2.5\baselineskip plus -1fil

\begin{IEEEbiography}[{\includegraphics[width=1in,height=1.25in,clip,keepaspectratio]{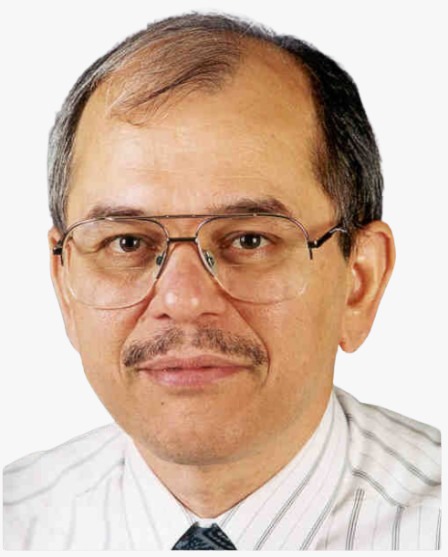}}]
{Kuldip K. Paliwal} received the B.S. degree from Agra University, India, in 1969, the M.S. degree from Aligarh Muslim University, India, in 1971, and the Ph.D. degree from Bombay University, India, in 1978. He has been active in speech and signal processing research since 1972 and has held research and academic positions at several institutions, including AT\&T Bell Laboratories, USA, the University of Keele, U.K., and Advanced Telecommunications Research Laboratories, Japan. Since 1993, he has been a Professor with Griffith University, Australia. His research interests include speech and image processing, pattern recognition, speech enhancement, speaker and face recognition, and neural networks. He has published more than 300 research articles and co-edited two books in speech and speaker recognition. He has served on several IEEE Signal Processing Society technical committees and editorial boards, including as Associate Editor of the \emph{IEEE Transactions on Speech and Audio Processing} and Editor-in-Chief of the \emph{Speech Communication} journal from 2005 to 2011. He received the IEEE Signal Processing Society Best (Senior) Paper Award in 1995 and is a Fellow of the Acoustical Society of India.

\end{IEEEbiography}

\vskip -2.5\baselineskip plus -1fil

\begin{IEEEbiography}[{\includegraphics[trim=250 300 250 50,width=1in,height=1.25in,clip,keepaspectratio]{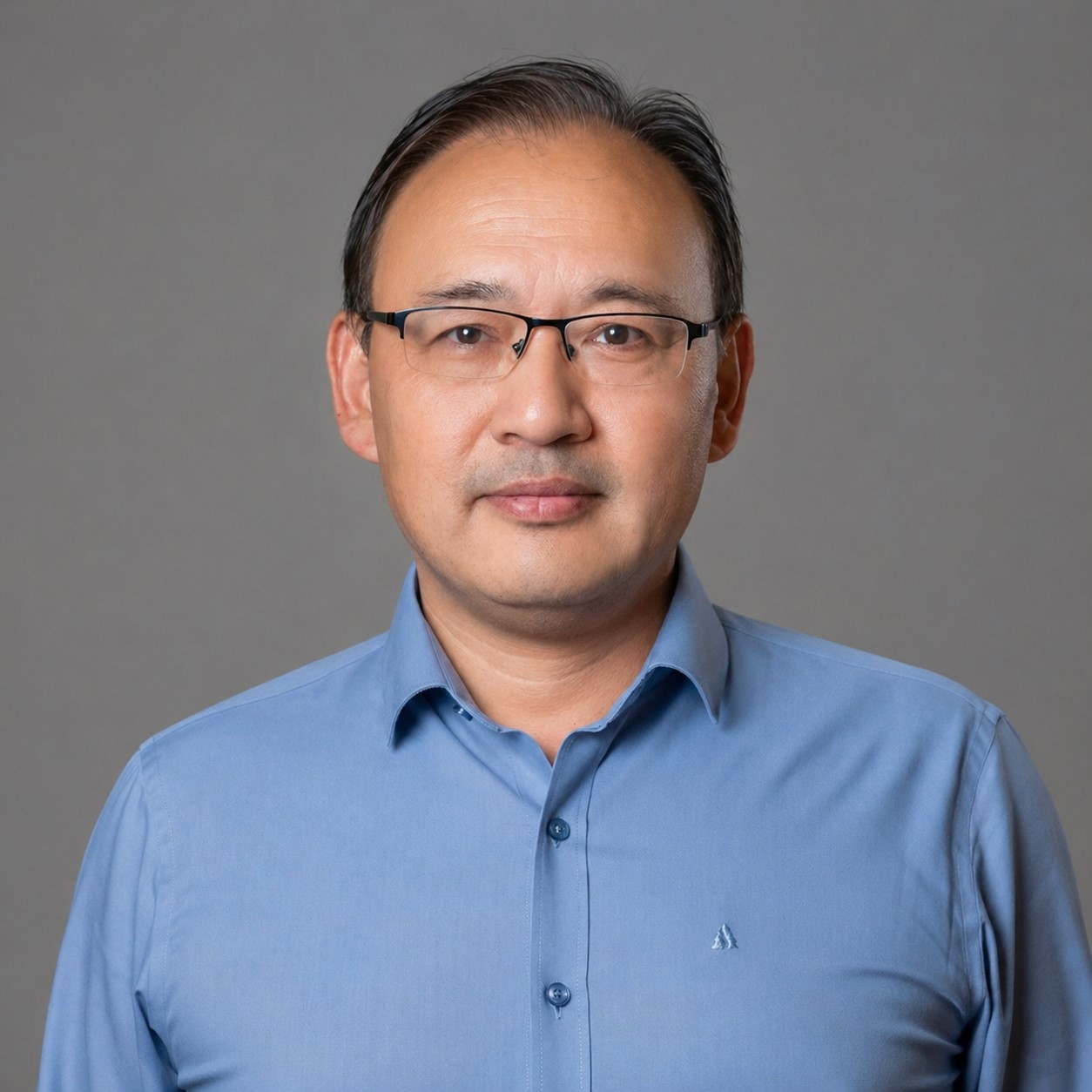}}]{Jun Zhou}
(Fellow, IEEE) received the B.S. degree in computer science and the B.E. degree in international business from the Nanjing University of Science and Technology, Nanjing, China, in 1996 and 1998, respectively, the M.S. degree in computer science from Concordia University, Montreal, QC, Canada, in 2002, and the Ph.D. degree from the University of Alberta, Edmonton, AB, Canada, in 2006. In June 2012, he joined the School of Information and Communication Technology, Griffith University, Nathan, QLD, Australia, where he is currently a Professor. Before this appointment, he was a Research Fellow in the Research School of Computer Science at the Australian National University, Canberra, ACT, Australia, and a Researcher at the Canberra Research Laboratory, NICTA, Canberra. His research interests include pattern recognition, computer vision, and spectral imaging with their applications in remote sensing and environmental informatics. Dr. Zhou is an Associate Editor of the IEEE Transactions on Multimedia, the IEEE Transactions on Geoscience and Remote Sensing, and the Pattern Recognition Journal.
\end{IEEEbiography}

\end{document}